\RequirePackage{fix-cm}

\documentclass[twocolumn]{svjour3}

\smartqed

\usepackage{graphicx}
\usepackage{epstopdf}
\usepackage{color}
\usepackage[square,numbers,sectionbib]{natbib}
\usepackage{hyperref}
\usepackage[all]{hypcap}
\usepackage{url}
\usepackage{listings}
\usepackage{floatflt}
\usepackage{tabu}
\usepackage[table]{xcolor}
\usepackage{mathtools}
\usepackage{booktabs}
\usepackage{amssymb}
\usepackage{changepage}

\newcommand{\devtext}[1]{} 

\begin{document}

\title{Named Entity Evolution Recognition on the Blogosphere\thanks{This work is partly funded by the European Commission under ARCOMEM (ICT 270239) and Alexandria (ERC 339233)}}

\titlerunning{Named Entity Evolution Recognition on the Blogosphere} 

\author{Helge Holzmann \and Nina Tahmasebi \and Thomas Risse}
\authorrunning{Holzmann, Tahmasebi, Risse}

\institute{Helge Holzmann \and Thomas Risse \at L3S Research Center, Appelstr. 9, 30167 Hannover, Germany, \email{\{holzmann, risse\}@L3S.de}
\and
Nina Tahmasebi \at Spr{\aa}kbanken, Department of Swedish, University of Gothenburg, Sweden, \email{nina.tahmasebi@gu.se}}

\date{Received: date / Accepted: date} 

\maketitle

\begin{abstract}
\begin{sloppypar}
Advancements in technology and culture lead to changes in our language. These changes create a gap between the language known by users and the language stored in digital archives. It affects user's possibility to firstly \textit{find} content and secondly \textit{interpret} that content. In previous work we introduced our approach for Named Entity Evolution Recognition~(NEER) in newspaper collections. Lately, increasing efforts in Web preservation lead to increased availability of Web archives covering longer time spans. However, language on the Web is more dynamic than in traditional media and many of the basic assumptions from the newspaper domain do not hold for Web data. In this paper we discuss the limitations of existing methodology for NEER. We approach these by adapting an existing NEER method to work on noisy data like the Web and the Blogosphere in particular. We develop novel filters that reduce the noise and make use of Semantic Web resources to obtain more information about terms. Our evaluation shows the potentials of the proposed approach.
\end{sloppypar}

\end{abstract}

\section{Introduction}

As time passes by, not just the world changes but also our language evolves. We invent new words, add or change meanings of existing words and change names of existing things. This results in a dynamic language that keeps up with our needs and provides us the possibility to express ourselves and describe the world around us. 

This process is fostered by the introduction of new technologies, especially the Web. It changes the way we express ourselves \cite{segerstad2002use}. In Social Media, for example blogs, everyone can publish content, discuss, comment, rate, and re-use content from anywhere with minimal effort. The constant availability of computers and mobile devices allows communicating with little effort, few restrictions, and increasing frequency. As there are no requirements for formal or correct language, authors can change their language usage dynamically. 

The resulting phenomenon is called \textbf{language evolution} (or \textbf{language change} in linguistics). For all contemporary use, language evolution is trivial as we are constantly made aware of the changes. At each point in time, we know the most current version of our language and, possibly, some older changes. However, our language does not carry a memory; words, expressions and meanings used in the past are forgotten over time. Thus, as users, we are limited when we want to find and interpret information about the past from content stored in digital archives. 

Awareness of language evolution is in particular important for searching tasks in archives due to the different ages of the involved texts and only a system that is aware of this knowledge can support information retrieval, for example by augmenting query terms. In the past, published and preserved content were stored in repositories like libraries and access was simplified with the help of librarians. These experts would read hundreds of books to help students, scholars or interested public to find relevant information expressed using any language, modern or old. 

Today, more and more effort and resources are spent digitizing and making available historical resources that were previously available only as physical hard copies, as well as gathering modern content. However, making the resources available to the users has little value in itself; the broad public cannot fully understand or utilize the content because the language used in the resources has changed, or will change, over time. The sheer volume of content prevents librarians to keep up and thus there are no experts to help us to find and interpret information. To fully utilize the efforts of digital libraries, this vast pool of content should be made semantically accessible and interpretable to the public. Modern words should be \textit{translated} into their historical counterparts and words should be represented with their past meanings and senses.

Language evolution is a broad area and covers many sub-classes like word sense evolution, term to term evolution, named entity evolution and spelling variations. In  this  paper,  we  will  focus  on  named  entity  evolution recognition (NEER). The task of NEER is to find name changes of entities over time, e.g. the former name of Pope Francis, in this case Jorge Mario Bergoglio. In \cite{Tahmasebi2012} we proposed an unsupervised method to find name changes without using external knowledge sources in newspaper archives. 

With the increasing number of Web archives being created, the language used on the Web looms large in these archives. \textit{Web language} is often closer to spoken language than to language used in traditional media~\cite{tpdl2012}. While the Web is getting older and Web archives are growing larger, keeping track of name changes will become as important as in traditional libraries. In this paper we present an adaption of the original NEER approach \cite{Tahmasebi2012} towards web language. For the evaluation we use two blog datasets that represent language on the Web in different intensities. We go beyond purely statistical methods by making use of the Web or the Semantic Web respectively and present a novel semantic filtering method. The filter helps to reduce erroneously detected name changes and improve the accuracy of the algorithm.

The next section shows an overview of different types of evolution and the corresponding problems caused. We show up the differences between digitized, historical content and archives with new content, e.g., Web archives. In Sect.~\ref{sec:NEER} we give an introduction to the original NEER approach and motivate the adaption by showing limitations of the method when applied on noisy data, e.g., from the Web. Sect.~\ref{sec:approach} presents our modified approach to NEER on the Web. We explain the additional noise reduction steps as well as the novel semantic filtering method, utilizing external resources from the Semantic Web. Sect.~\ref{sec:experiments} contains the details of our evaluation, including a description of the dataset, the test set and the parameters used. The results are discussed in Sect.~\ref{sec:discussions}. In Sect.~\ref{sec:sota} we provide a review of current methods for detecting named entity evolution as well as related research that is fundamental for our approach. Finally, in Sect.~\ref{sec:conclusions} we conclude and discuss future directions to make digital archives semantically accessible and interpretable, thus ensuring useful archives also for the future.

\section{Language Evolution}

\begin{sloppypar}
There are two major problems that we face when searching for information in long-term archives; firstly \textit{finding} content and secondly, \textit{interpreting} that content.
When things, locations and people have different names in the archives than those we are familiar with, we cannot find relevant documents by means of simple string matching techniques. The strings matching the modern name will not correspond to the strings matching the names stored in the archive. The resulting phenomenon is called \textbf{named entity evolution} and can be illustrated with the following: 
\begin{quote}
``The Germans are brought nearer to \underline{Stalingrad} and the command of the lower Volga.''
\end{quote} 
The quote was published on July 18, 1942 in The Times \citep{Times-Stalingrad} and refers to the Russian city that often figures in the context of World War II. In reference to World War II people speak of \textit{the city of Stalingrad} or the \textit{Battle of Stalingrad}, however, the city cannot be found on a modern map. In 1961, \textit{Stalingrad} was renamed to \textit{Volgograd} and has since been replaced on maps and in modern resources. Not knowing of this change leads to several problems; 1. knowing only about \textit{Volgograd} means that the history of the city becomes inaccessible because documents that describe its history only contain the name \textit{Stalingrad}. 2. knowing only about \textit{Stalingrad} makes is difficult to find information about the current state and location of the city\footnote{Similar problems arise due to spelling variations that are not covered here.}.
\end{sloppypar}

The second problem that we face is related to interpretation of content; words and expressions reflect our culture and evolve over time. Without explicit knowledge about the changes we risk placing modern meanings on these expressions which lead to wrong interpretations. This phenomenon is called \textbf{word sense evolution} and can be illustrated with the following: 
\begin{quote}

``Sestini's benefit last night at the Opera-House was overflowing with the fashionable and \underline{gay}.''
\end{quote}
The quote was published in April 27, 1787 in The Times \citep{Times-Opera}. When read today, the word {\em gay} will most likely be interpreted as {\em homosexual}. However, this sense of the word was not introduced until early 20th century and instead, in this context, the word should be interpreted with the sense of {\em happy}. 

\begin{sloppypar}
Language evolution also occurs in shorter time spans; modern examples of named entity evolution include company names (\textit{Andersen Consulting}$\longrightarrow$\textit{Accenture}) and Popes (\textit{Jorge Mario Bergoglio}$\longrightarrow$\textit{Pope Francis}). Modern examples of word sense evolution include words like \textit{Windows} or \textit{surfing} with new meanings in the past decades. 
\end{sloppypar}

In addition, there are many words and concepts that appear and stay in our vocabulary for a short time period, like  \textit{smartphone face}, \textit{cli-fi} and \textit{catfishing}~\footnote{http://www.wordspy.com/} that are examples of words that have not made it into e.g., Oxford English Dictionary, and are unlikely to ever do so. 

\subsection{Types of Evolution}

\begin{figure}[bh!]
\includegraphics[width=\columnwidth]{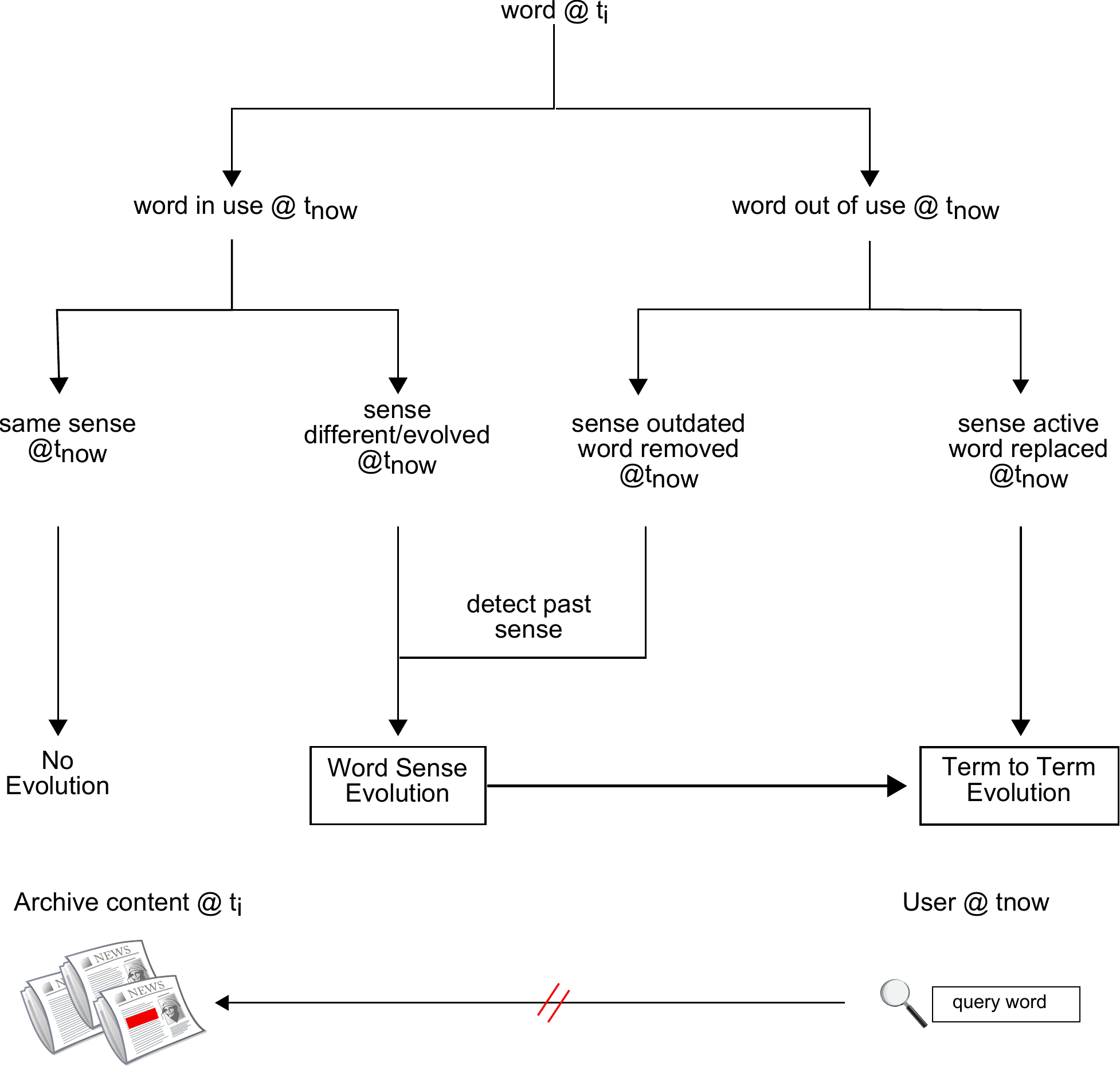}
\caption[Evolution of Words]{Diagram of Word Evolution}
\label{fig:WordEvoDiagram}
\end{figure}

Formally, the problems caused by language evolution (illustrated in Fig.~\ref{fig:WordEvoDiagram}) can be described with the following: Assume a digital archive where each document $d_i$ in the archive is written at some time $t_i$ prior to current time $t_{now}$. The larger the time gap is between $t_i$ and $t_{now}$, the more likely it is that current language has experienced evolution compared to the language used in document $d_i$. For each word $w$ and its intended sense $s_w$ at time $t_i$ in $d_i$ there are two possibilities; 1. The word can still be in use at time $t_{now}$ and 2. The word can be out of use (outdated) at time $t_{now}$. 

Each of the above options opens up a range of possibilities that correspond to different types of language evolution that affect finding and interpreting in digital archives.

\underline{\bf Word $w$ at time $t_i$ in use at $t_{now}$}

\textbf{No Evolution:} The word is in use at time $t_{now}$ and has the \textit{same sense} $s_w$ and thus there has been no evolution for the word. The word and its sense are stable in the time interval [$t_i,t_{now}$] and no action is necessary to understand the meaning of the word or to find content. 

\textbf{Word Sense Evolution:} The word is still in use at time $t_{now}$ but with a \textit{diff\-erent sense} $s'_w$. The meaning of the word has changed, either to a completely new sense or to a sense that can be seen as an evolution of the sense at time $t_i$. The change occurred at some point in the interval ($t_i,t_{now}$). We consider this to be the manifestation of word sense evolution. 
 
\underline{\bf Word $w$ from $t_i$ out of use at $t_{now}$}

{\bf Word Sense Evolution - Outdated Sense: } The word is out of use because the word sense is outdated and the word is no longer needed in the language. This can follow as a consequence of, among others, technology, disease or occupations that are no longer present in our society. The word $w$ as well as the associated word sense $s_w$ have become outdated during the interval ($t_i,t_{now}$). To be able to interpret the word in a document from time $t_i$ it becomes necessary to detect the active sense $s_w$ at time $t_i$. Because it is necessary to recover a word sense that is not available at time $t_{now}$ we consider this to be a case of word sense evolution. 

{\bf Term to Term Evolution:} The word $w$ is outdated but the sense $s_w$ is still active. Therefore, there must be another word $w'$ with the same sense $s_w$ that has replaced the word $w$. That means, different words, in this case $w$ and $w'$, are used as a representation for the sense $s_w$ and the shift is made somewhere in the time interval ($t_i, t_{now}$). We consider this to be term to term evolution where the same sense (or entity) is being represented by two different words. If the word $w$ represents an entity, we consider it to be named entity evolution. 

In addition to the above types of evolution, there are also \textit{spelling variations} that can affect digital archives; historical variations with different spellings for the same word or modern variations in the form of e.g., abbreviations and symbols. Spelling variations are not considered in this paper.

\subsection{Historical vs. Modern Data -- Old vs. New Content} \label{sec:histData}
When working with language evolution from a computational point of view there are two main perspectives available. The first considers today as the point of reference and searches for all types of language evolution that has occurred until today. In this perspective the language that we have today is considered as common knowledge and understanding past language and knowledge is the primary goal. 

In the second perspective the goal is to prepare today's language and knowledge for interpretation in the future. We monitor the language for changes and incrementally map each change to what we know today. We can assume that knowledge banks and language resources are available and most new changes are added to the resources.

There are however several limitations with modern language as well. The first limitation is noisy data being published on the Web. With increasing amounts of user generated text and lack of editorial control, there are increasing problems with grammars, misspellings, abbreviations, etc. To which level this can be considered as real noise like with OCR errors is debatable, however, it is clear that this noise reduces the efficiency of tools and algorithms available today. This in turn limits the quality of evolution detection as we depend on existing tools and their efficiency. The second limitation is the restricted nature of resources like Wikipedia. As with dictionaries, Wikipedia does not cover all entities, events and words that exist. Instead, much is left out or only mentioned briefly which limits to which extent we can depend exclusively on these resources.

In order to avoid that future generations face the same problems as we have to face, we need to start thinking about these problems already now. In particular for Web archives that are continuously created and updated, with ephemeral words, expressions and concepts. Otherwise we risk to render a large portion of our archives semantically inaccessible and cannot utilize the great power of crowd sourcing.

\subsection{Challenges on Web Data}

\citet{tpdl2012} showed that language in blogs behaves differently than traditional written language. We take this representatively for the language used on the Web, referred to as Web language. This language is a mixture of different local and global trends from all users on the Web as everyone can contribute their ideas and thoughts. Accordingly, Web language compared to typical written language is more dynamic and closer to spoken language. This has been reinforced even more by the introduction of Social Media, like blogs or social networks, and the increasing ubiquity of computers and mobile devices. These technologies lower the restrictions and limitations for authors with no professional background. Everyone can publish content with minimal effort as well as discuss, comment, rate, and re-use content from anywhere. This leads to a less formal and correct language on the Web, with unconventional spellings and a colloquial terminology.

Furthermore, in our experiments we could observe that Web data compared to texts in traditional media leads to more entities in the result set, even though the entities are not of direct relevance. Those entities are, among others, alternative products, neighboring countries or predecessors in office. Our hypothesis is that this behavior is caused by the structure of the Web. As the Web is a graph of interlinked texts, the authors are more encouraged to create hyperlinks to related articles than in traditional media. For example, while newspaper simply report about a new product and keep it short due to lack of space, Blogs typically refer to similar products in order to link to their related articles.

For these reasons, texts on the Web need to be treated differently than, for instance, texts in traditional newspapers. NEER on the Web requires higher robustness against the dynamics of Web language. More advanced filtering techniques are required to filter additional, colloquial and informal terms. This noise must be prevented from making its way to the context of a term and thus being considered as potential co-references. Therefore, a specialized NEER approach for the Web is needed.

In addition to these challenges, the Web opens the door to new opportunities. The Semantic Web provides additional data that can support finding co-references of a term. For example, the information that \textit{Vatican} is not a person can be used to identify it as a wrong co-reference for \textit{Pope Benedict XVI}, although it might be detected as a name change as the terms co-occur very frequently. One might argue that this knowledge is redundant as knowledge bases already know about the co-references of a term, however, no knowledge base can cover all entities. Moreover, the information about co-references need to be present as an explicit property in knowledge bases to be used by computers. With NEER, on the other hand, also implicitly mentioned temporal co-references in texts can be recognized. For example, \textit{Czechoslovakia} does not have a property on DBpedia that explicitly reflects its name evolution by the split into \textit{Czech Republic} and \textit{Slovakia}, even though these terms are temporal co-references of each other. Comparing their properties, however, reveals the likelihood of evolution. Therefore, especially when working with data from the Web, it is reasonable to incorporate semantic information of terms to filter out noise and erroneously detected co-references.

\subsection{Problem and Motivation}
\label{sec:problem}

The definition above shows the complexity of language evolution in general. As we focus on Named Entity Evolution Recognition (NEER) we can formulate the problem more strictly.

\begin{sloppypar}
{\bf Input} As input for NEER we consider a tuple $(q, DS, ad)$ consisting of a query term $q$, a dataset $DS$ and additional data $ad$. The dataset $DS = (D, S)$ is a pair containing documents $D$ and a set of sources $S$ (i.e., different newspapers, websites, etc.). Every document $d \in D$ is a triple with content $c_d$ (i.e., the text), the time $t_d$ when it was published and a source $s_d \in S$ where it was published:
\[d = (c_d,\ t_d,\ s_d)\]
\end{sloppypar}

Every text $c$ contains a set of terms $W_c$. $W_{DS}$ denotes the set of all terms present in the dataset $DS$:
\[W_{DS} = \bigcup_{c \in \{c_d | d \in D\}} W_c\]

For a given query term $q$ and a dataset $DS$ the task of NEER is to detect all name changes of $q$ that are reported in the dataset as well as all variations of $q$ that are used in the dataset to refer to $q$. Additional data $ad$ supports this task by providing extra information about entities or terms. In the extension presented in this article, the additional data consists of semantic information provided by an external knowledge base (i.e., DBpedia).

\begin{sloppypar}
{\bf Output} A test set defines the name changes that we expect to be found by NEER. The goal is to find a result set that matches the test set in terms of completeness as well as accuracy. That means, all name changes of the query term as defined by the test set should be found, but no other terms. Erroneously detected names are referred to as \textit{false positives}.
\end{sloppypar}

\begin{sloppypar}
A test set $T$ is defined as a set of expected test tuples $\mathit{test}_q \in T$, one for each query term $q$. In addition to the query term, $\mathit{test}_q$ contains a set of expected names $E_{q}$ for $q$ as well as the change periods $P_{q}$ of the entity referred to by $q$:
\[\mathit{test}_q = (q,\ E_q,\ P_q)\]
\end{sloppypar}

\begin{sloppypar}
For example, the set of expected name changes $E_q$ for the query $q =$ \textit{Barack Obama} contains \textit{Senator Obama}. \textit{Barack Hussein Obama} and \textit{President Obama}. The corresponding set of change periods $P_q = \{2004, 2008\}$ consists of the years when \textit{Barack Obama} became senator and president.
\end{sloppypar}

A result set $R_q$ consists of the name changes detected by NEER for the query term $q$ utilizing the change periods $P_q$. To determine the completeness and accuracy of $R_q$ we use the recall and precision metrics:
\[recall_q = \frac{|R_q \cap E_q|}{|E_q|}\]
\[precision_q = \frac{|R_q \cap E_q|}{|R_q|}\]

The goal of a NEER algorithm is to maximize these metrics. However, we consider high precision more important than high recall. The reason for this is, thinking of NEER as support for a search engine, a utility to improve the search result. With high precision but low recall the search result is not improved very much but the results are also not worsened. A lower precision, by contrast, means that the NEER results contain more false positives. This may lead to a worse search result, despite the achieved recall.

\section{NEER}
\label{sec:NEER}

\begin{figure*}[t]
\centering
\includegraphics[width=\textwidth]{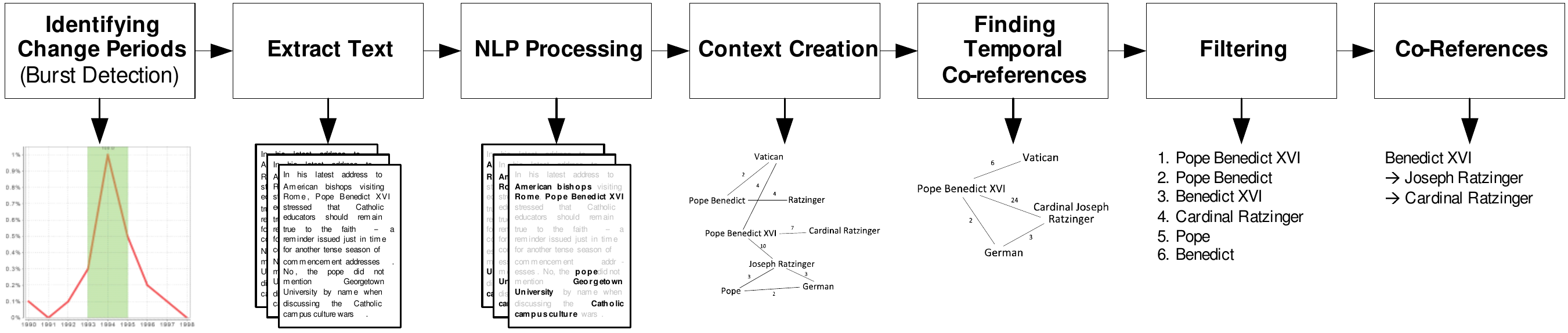}
\caption{BaseNEER pipeline for temporal co-reference detection\cite{Tahmasebi2012}.}
\label{fig:NEER_workflow}
\end{figure*}

\begin{sloppypar}
Named entity evolution recognition addresses the problem of automatically detecting name changes for entities. The method presented by \cite{Tahmasebi2012} is an unsupervised approach for NEER making use of high quality newspaper datasets (i.e., New York Times Annotated Corpus). In this paper, the approach is used as a foundation as well as the baseline for our adaption. Therefore, in the following we reference it as \textbf{BaseNEER}, while we call our approach \textbf{BlogNEER}. BlogNEER is an extension of BaseNEER for reducing noise and incorporating external resources in order to apply NEER to more noisy data like on the Web and on the Blogosphere in particular. This section gives an overview of BaseNEER, its limitations on Web data and provides definitions that we use throughout the article.
\end{sloppypar}

\subsection{Definitions} 
\label{sec:baseneer_definitions}

The basic terminology that is needed for NEER has been defined by \citet{Tahmasebi2012}. For this work we use the same definitions as given below and extend them by introducing new specific terminology needed for BlogNEER in Sect.~\ref{sec:webneer_definitions}.

A \textbf{term}, typically denoted as $w_i$, is considered to be a single or multi-word lexical representation of an entity $e_i$ at time $t_i$. All terms that are related to $w_i$ at time $t_i$ are logically grouped in the \textbf{context} $C_{w_i}$. Similar to \citet{Berberich2009}, a context of $w_i$ is considered to be the most frequent terms that co-occur within a certain distance to $w_i$. 

A \textbf{change period} of the entity $e_i$ is a period of time likely to surround an event that yielded a new name of $e_i$. Different names referring to the same entity $e_i$ are denoted as \textbf{temporal co-references} of $e_i$. Temporal co-references contain names that are used at the same time as well as at different times. For the sake of simplicity we use the terms  \textit{co-reference} and \textit{temporal co-reference} interchangeably. 
 
Co-references can be classified as direct and indirect. BaseNEER considers a \textbf{direct temporal co-reference} of $e_i$ to be a co-references with some lexical overlap with $e_i$. For \textit{BlogNEER} we generalize this definition as we consider a \textbf{direct temporal co-reference} of $e_i$ to be a co-reference that has been derived by incorporating only lexical features. By contrast, on computing \textbf{indirect temporal co-references} additional features, like co-occurrence relations, can be utilized as well. As an example, \textit{President Obama}, \textit{Barack Obama} and \textit{President} may be considered as direct co-references, even though \textit{Barack Obama} lacks lexical overlap with \textit{President}. The lexical connection is given through \textit{President Obama}.

\begin{sloppypar}
All direct temporal co-references that have been found by BaseNEER for an entity $e_i$ are grouped into a \textbf{temporal co-reference class}. A co-reference class is represented by the most frequent member $r$ of the class, called a class representative. We denote a temporal co-reference class as $coref_r\ \{w_1,\ w_2\ ,\ \dots\}$. For instance, the co-reference class containing the terms \textit{Joseph Ratzinger, Cardinal Ratzinger, Cardinal Joseph Ratzinger, \dots} with representative \textit{Joseph Ratzinger} is denoted as $coref_{Joseph\ Ratzinger}$ \textit{\{Joseph Ratzinger, Cardinal Ratzinger, Cardinal Joseph Ratzinger, \dots\}}. For BlogNEER this class is defined slightly different and called sub-term classes with a similar notation (s. Sect. 4.1).
\end{sloppypar}

\subsection{Overview of BaseNEER}
\label{sec:BaseNEER}

The major steps of the BaseNEER approach are depicted in Fig.~\ref{fig:NEER_workflow}. BaseNEER utilizes change periods for detecting name evolutions of entities. 

The first step of the BaseNEER pipeline is the burst detection for identifying change periods. In this step high frequency bursts of an entity are identified in the used dataset. The year around such a burst is considered a change period.

Documents from such a change period are regarded for collecting co-reference candidates of the corresponding entity. All terms that can be considered names (i.e., named entities) are extracted from the documents. These terms are used to build up a graph that represents co-occurrences among the terms in the considered texts. This graph defines the context of the entity. 

The following step of finding temporal co-references is the core of BaseNEER. Based on the derived context graph, four rules are applied to find direct co-references among the extracted terms. These are merged to co-reference classes as follows:
\begin{enumerate}
\item \textit{Prefix/suffix rule}: Terms with the same prefix/suffix are merged (e.g., Pope Benedict and Benedict).
\item \textit{Sub-term rule}: Terms with all words of one term are contained in the other term are merged (e.g., Cardinal Joseph Ratzinger and Cardinal Ratzinger).
\item \textit{Prolong rule}: Terms having an overlap are merged into a longer term (e.g., Pope John Paul and John Paul II are merged to Pope John Paul II).
\item \textit{Soft sub-term rule}: Terms with similar frequency are merged as in rule 2, but regardless of the order of the words. 
\end{enumerate}

Subsequent to merging terms, the graphs are consolidated by means of the co-references classes. Thus, only representatives remain as nodes while the edges to other terms in a class are connected to their representatives. 

Afterwards, filtering methods are used to filter out false co-references that do not refer to the query term. For this purpose, statistical as well as machine learning (ML) based filters were introduced. A comparison of the methods revealed their strengths and weaknesses with respect to precision and recall. The ML approach performed best with noticeable precision and recall of more than 90\% and 80\% respectively. Even though it is possible to reach a high accuracy with NEER + ML, training the needed ML classifier requires manual labeling. Therefore, we do not consider this filter as completely unsupervised and do not use it as a baseline for the evaluation.

\section{BlogNEER Approach}
\label{sec:approach}

\begin{figure*}[t]
\centering
\includegraphics[width=\textwidth]{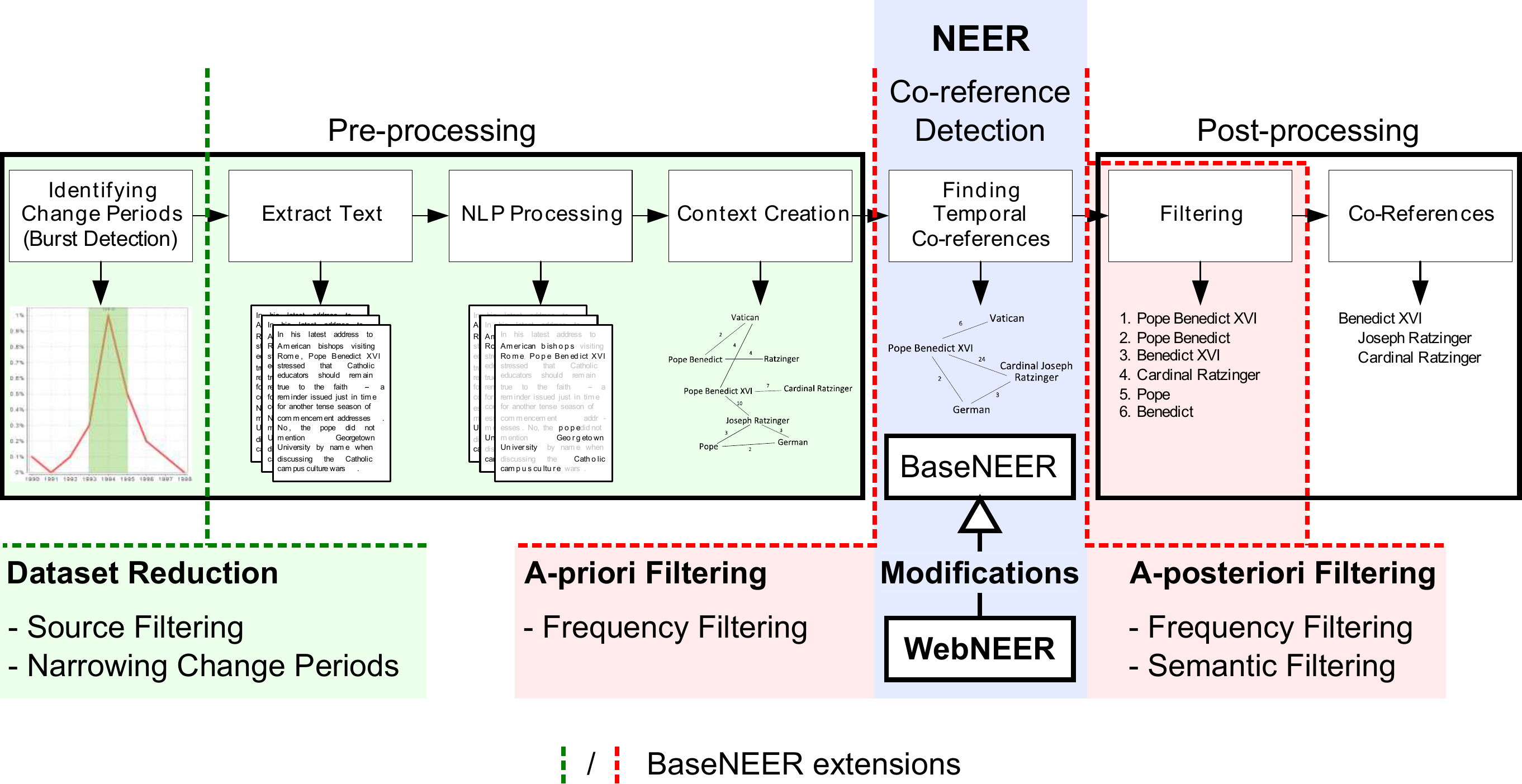}
\caption{BlogNEER extensions on the BaseNEER pipeline.}
\label{fig:filtering_hooks}
\end{figure*}

As a consequence of a more dynamic language, fewer restrictions and lower requirements in terms of quality, texts on the Web contain a larger variety of terms than higher quality datasets, such as newspapers. These terms cause larger contexts (i.e., larger amounts of unique co-occurring terms) and lead to more co-references derived by BaseNEER when applied to Web data. From the NEER perspective most of these terms are noise and therefore, lower the precision.

\begin{sloppypar}
The challenge is to filter out the noise while keeping the true co-references of a query term. We tackle this by extending the BaseNEER pipeline (s. Fig.~\ref{fig:NEER_workflow}) to be less noise prone. The extensions are depicted in Fig.~\ref{fig:filtering_hooks}. A new dataset reduction step as well as additional a-priori and a-posteriori filters reduce the terms in the query term's context. Additionally, we modify the co-reference detection step of BaseNEER to be more resistant against noise. A novel semantic filter incorporates semantic resources of the terms in order to identify those terms that do not refer to the same entity as the query term. 
\end{sloppypar}

\subsection{Definitions}
\label{sec:webneer_definitions}

For the extensions of BaseNEER, that is BlogNEER, we need some additional, specific definitions. In the co-reference detection process of BaseNEER one of the core elements is the co-reference class. In BlogNEER this is substituted by sub-term classes. We define a \textbf{sub-term} of term $w$ to be a complete single token term that is included in $w$. For example, the sub-terms of \textit{President Obama} are \textit{President} and \textit{Obama}. Accordingly, we define a \textbf{super-term} of $w$ a term that contains all sub-terms of $w$. Hence, \textit{President Barack Obama} is a super-term of \textit{President Obama}. Additionally, every term is a super-term of itself. With these definitions we consider a sub-term class $\mathit{sub}_w$ to be a group of terms that only consists of sub-terms of $w$, or in other words, $w$ is a super-term of all of these terms. The super-term class of $w$, denoted as $\mathit{super}_w$, contains all super-terms of $w$ and therefore, all terms that include $w$ in their sub-term classes. Sub-term classes, instead of co-reference classes, are detected by BlogNEER during co-reference detection. In Sect.~\ref{sec:NEER_modifications} we explain how these changes can help to reduce noise.

\begin{sloppypar}
Similar to the notation of a co-reference class we denote a sub-term class of the representative $r$ as $sub_r\ \{r,\ w_1,\ w_2\ ,\ \dots\}$. For example, the sub-term class of \textit{Union of Myanmar}, containing the sub-terms \textit{Union}, \textit{Myanmar} and is denoted as $\mathit{sub}_\mathit{Union\ of\ Myanmar}$ \textit{\{Union of Myanmar, Union, Myanmar\}}. In addition, for better readability in some examples we use the notation ``Representative [$w_1,\ w_2,\ \dots$]'' (with $w_i \neq r$ for $i=1,2,...$), like ``Union of Myanmar [Union, Myanmar]''.
\end{sloppypar}

Based on the detected sub-term classes we compute direct and indirect co-references of a term, as defined for BaseNEER (s. Sect.~\ref{sec:baseneer_definitions}). These are derived from a context graph. In the beginning the graph consists of terms as nodes and edges represented by co-occurrences among the terms. After computing the sub-term classes, these supplant the terms as nodes. Also edges are consolidated and the sub-term classes are connected with edges based on the co-occurrences among the terms in the classes. In order to refer to the connected terms or classes of a node $n$ we introduce the function $\mathit{related}(n)$. It yields a set of terms or sub-term classes, depending on the type (term or sub-term class) of $n$ (E.g., for the sub-term class $A$, connected to the sub-term classes $B$ and $C$, $\mathit{related}(A) = \{B, C\}$). Additionally, since temporal the co-references are not directly specified by the classes anymore, we introduce three functions that return the co-references of $w$, one for direct, one for indirect and one for the union of direct and indirect co-references:

\begin{itemize}
\item $\mathit{direct\_corefs}(w)$
\item $\mathit{indirect\_corefs}(w)$
\item $\mathit{corefs}(w) = \mathit{direct\_corefs}(w)\ \cup\ \mathit{indirect\_corefs}(w)$
\end{itemize}

Each of these functions specifies a sets of terms. Before all filters have been applied and the sets can be considered final, we call these terms \textbf{candidates}. A term in $\mathit{corefs}(w)$ is a direct or indirect co-reference candidate of $w$ until it is filtered out or determined to be a temporal co-reference by BlogNEER.

\begin{sloppypar}
In additions to definitions and notations for the NEER process, we need to define terminology specific for the Web. The most characteristic elements of the Web are websites. While the data that BaseNEER operates on is a set of documents from one newspaper source, the documents on the Web originate from different websites. We call each of these websites a \textbf{source}. A source may be a traditional static website, a wiki, a blog or a social network stream. Each source consists of multiple items. Those items may be subpages of a website or wiki, blog articles or posts in a stream. In order to have one consistent notation, we refer to any kind of item in a source as a \textbf{document}. If a term is included in more than one document, these documents might have been published all in the same source or in different sources. We denote the number of the documents a term occurs in as \textbf{document frequency}. Accordingly, the \textbf{source frequency} of a term $w$ is the number of sources with documents that contain $w$.
\end{sloppypar}

Naturally, not all terms that occur in a dataset are actually relevant for NEER with respect to a given query term $q$. These irrelevant terms can worsen the result when they make their way into the NEER process. Therefore, we refer to them as \textbf{noise}. Noise that is considered a candidate or has been taken into the final result set is called a \textbf{false positive}. Noisy terms are not necessarily misspelled or colloquial, but also terms that often co-occur with $q$, for example descriptive terms or frequently co-occurring terms, like neighboring countries if $q$ is a country or competitors if $q$ is a company. These are often detected as co-references but should be considered noise. We denote these terms as \textbf{complementary terms}.

In order to filter out noise and reduce the number of false positives, BlogNEER incorporates an external knowledge source, called a \textbf{knowledge base} (i.e., DBpedia). A knowledge base consists of entries with information about terms or entities respectively. These entries are called \textbf{semantic resources}. The information in a semantic resource is organized as key-value-pairs with properties and values that describe the corresponding entity. We call these \textbf{semantic properties} of a term or entity respectively. On DBpedia a resource is the structured representation of a Wikipedia page. The semantic properties are automatically (or in some cases manually) extracted from the content of that page as described by \citet{Bizer2009}.

\subsection{BlogNEER Overview}

\begin{figure*}[t]
\centering
\includegraphics[width=\textwidth]{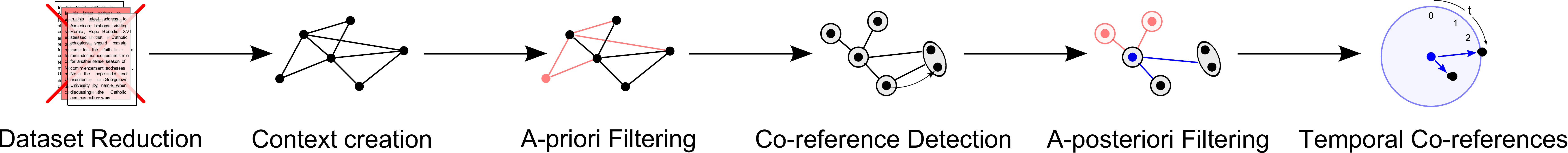}
\caption{BlogNEER workflow.}
\label{fig:webneer_pipeline}
\end{figure*}

BlogNEER extends the BaseNEER approach by adding additional filters and modifying the original method to be more robust to noise (s. Fig.~\ref{fig:filtering_hooks}). Fig.~\ref{fig:webneer_pipeline} depicts the new workflow of BlogNEER.

In order to deal with the noise we are facing on the Web, we introduced an additional pre-processing step, namely dataset reduction, as well as an a-priori frequency filter to reduce the large data we find on the Web with respect to the query. This reduces the amount of noise that makes its way to the co-reference detection. Afterwards, we apply two a-posteriori filters that filter out erroneously detected co-reference candidates. Another frequency filter choses only highly frequent candidates by means of the consolidated and emphasized frequencies after the co-reference detection. Finally, a semantic filter eliminates false positive results by incorporating semantic data from a knowledge base. This allows us to compare terms not just on lexical but also on a semantic level. Therefore, we can tell terms apart that refer to different entities if they have certain different semantic properties. For example, we can filter out \textit{Vatican} as co-reference for \textit{Pope Benedict} by identifying the first as a place and the latter as a person.

\subsection{Dataset Reduction}
\label{sec:dataset_reduction}

The dataset reduction aims to focus NEER on the relevant documents with respect to the query term and its change period. The sub-terms of a query may be ambiguous in different domains, like \textit{President}. However, for the purpose of NEER we are only interested in document from the domain of the entire query term (e.g., \textit{President Obama}). Having irrelevant documents in consideration can lead to more noise. With dataset reduction we approach this issue by filtering out sources from irrelevant domains and narrow the change period to concentrate on the actual name change event.

\subsubsection{Source Filtering}

\begin{sloppypar}
The Web consists of many different data sources (i.e., static websites, blogs, etc.), all consisting of texts about different topics from several domains. Thus, as we run BlogNEER on a subset of the Web (i.e., two blog datasets in our evaluation), we have to deal with documents from different domains, too. Out of these documents we select those that were published during a certain time period and contain at least one sub-term of the full query term. For example, for the query term \textit{President Obama}, we consider all documents from our dataset that include \textit{President} or \textit{Obama} during a specified period. However, the term \textit{President} for example is pretty ambiguous. There is the President of the United States that we are interested in, but also many other presidents from companies, sport clubs, etc.. Hence, it is not just a term in the political domain and we would probably find many erroneous terms by querying for \textit{President} in all data sources.
\end{sloppypar}

Therefore, we try to restrict our dataset to documents that are actually reporting about the term we are interested in. However, those are not only documents containing \textit{President Obama} entirely. Some documents might refer to him just as \textit{President} or \textit{Barack Obama}.

\begin{figure*}[t]
\centering
\includegraphics[width=\textwidth]{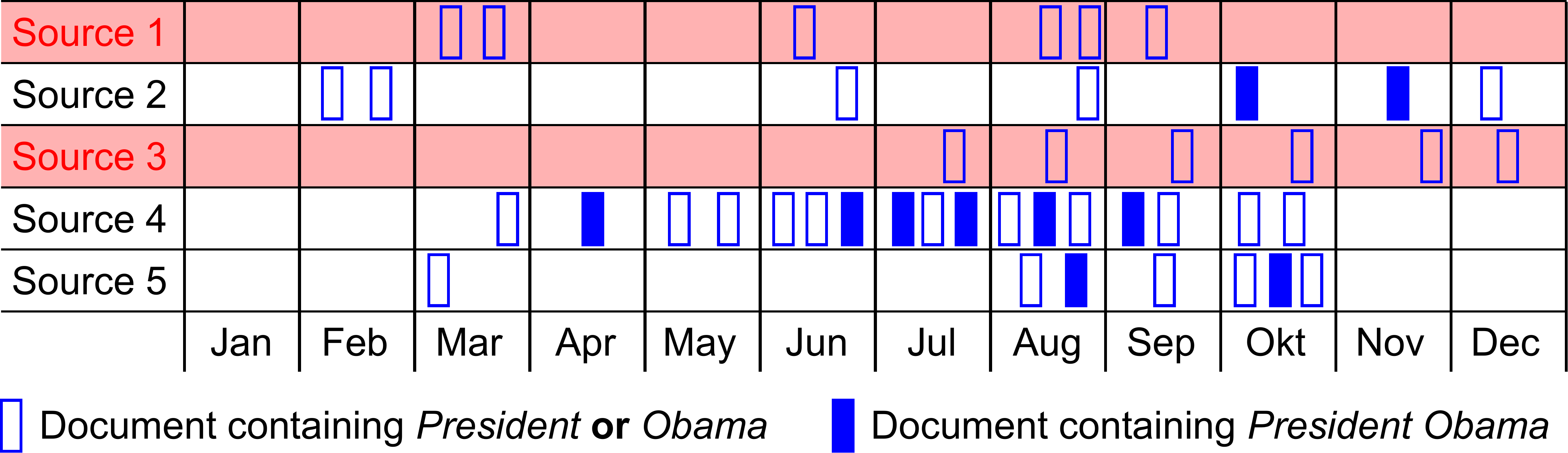}
\caption{Removing documents from sources that do not mention the complete query term. All documents of source 1 and source 3 are removed from our dataset for the query \textit{President Obama}, as no document from these sources contains the sub-terms \textit{President} or \textit{Obama}.}
\label{fig:source_filtering}
\end{figure*}

\begin{sloppypar}
A less restrictive selections would be to select only documents from sources of a certain domain. For \textit{President Obama} those are most likely documents from sources of the political domain. As this information rarely is available on the Web, we consider all sources containing the query term as a whole in a least one document during the specified period to be considered a source of the corresponding domain. In our example, we only consider the documents from sources that contain \textit{President Obama} during the specified change period in at least one document, as shown in Fig.~\ref{fig:source_filtering}. We filter out all documents from the other sources. Next the documents from the remaining sources are queried for the query sub-terms (i.e., \textit{President} or \textit{Obama}) and the retrieved documents are used in the further NEER process.
\end{sloppypar}

Formally, based on the definitions in Section~\ref{sec:problem}, this means out of all sources $S$ available in our dataset $DS$ we keep only those with at least one document $d \in D$ containing the entire query $q$ during the change period under consideration $p \in P_q$. Hence, our filtered source collection $S_f$ consists only of sources from these documents: $S_f = \{s_d | d \in D, q \in c_d, t_d \in p_q\}$. Accordingly, we only consider documents from these sources: $D_f = \{d \in D | s_d \in S_f\}$. The filtered dataset, which is considered in the further processing of query $q$ during change period $p$, is defined as $DS_{q,p} = (D_f, S_f)$.

\subsubsection{Narrowing Change Periods}

With BaseNEER the concept of change periods was introduced. Change periods are time periods likely to cover a name change of the corresponding entity. An entity can have one or multiple change periods. In BaseNEER, the year around a burst of the query term in the dataset is considered a change period. However, often the documents containing the query term do not occur during the full year. The actual name change in such a period could only be at the end of the year and there might be only one month around the event with documents reporting about it. As an example consider the election of \textit{President Obama}. A dataset consisting of sports Blogs might not contain \textit{President Obama} at other times at all. However, as the articles report about a sports presidents all year, a sub-term match for \textit{President} would contribute a lot of noise to the NEER process. To avoid this scenario, in BlogNEER we narrow the periods to a shorter time period during the full year when the documents actually contain the full query term.

\begin{figure*}[t]
\centering
\includegraphics[width=\textwidth]{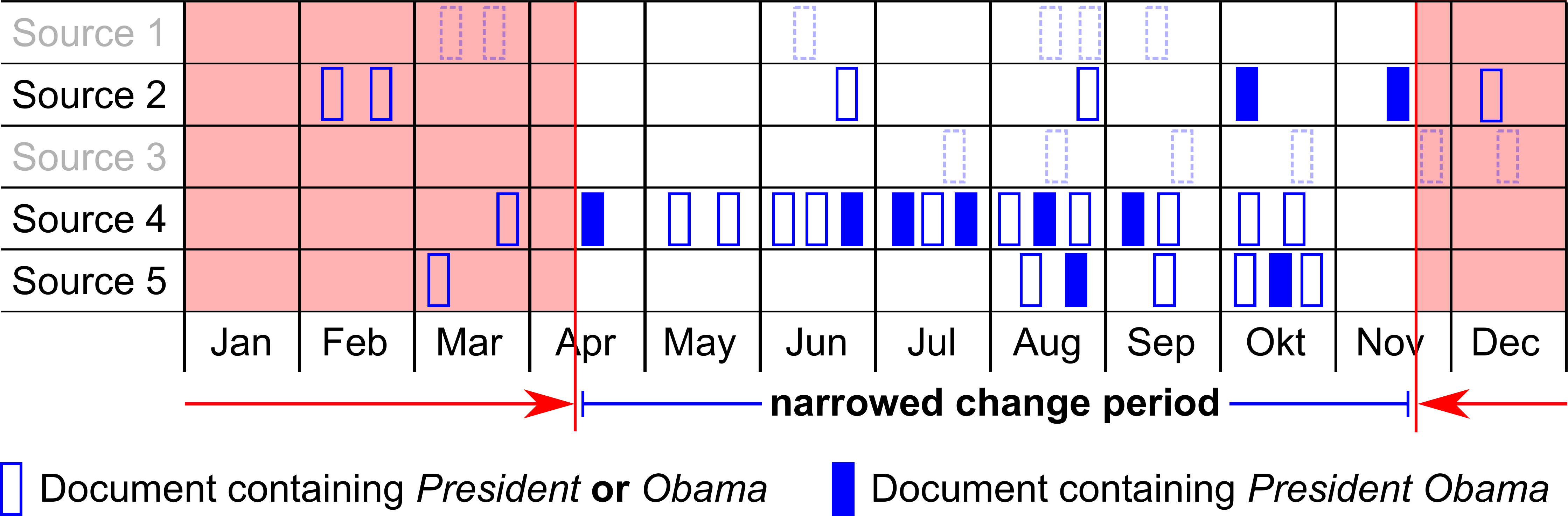}
\caption{Narrowing the change period to the time frame with documents containing the complete query term.}
\label{fig:change_periods}
\end{figure*}

\begin{sloppypar}
As Web datasets usually contain multiple sources, we search in all of them for documents that were published during the change period $p$ and contain the query term $q$. For each source we determine the earliest and latest publishing date of the found documents. As Fig.~\ref{fig:change_periods} depicts, the minimum ($minDate(q,p)$) and maximum ($maxDate(q,p)$) of these dates are used (regardless of which source they originate) to refine the change period. The narrowed period starts at $minDate(q,p)$ and ends at $maxDate(q,p)$. This period will be considered the new change period $p$ for the further process.
\end{sloppypar}

To attain these values a query for the full term $q$ in the considered period $p$ is done during the pre-processing phase. Out the returned collection of documents $D_{pq} \subset D \in DS$, the earliest and latest documents are considered as the desired boundaries:\\$minDate(q,p) = \min\{t_d | d \in D_{p,q}\}$\\$maxDate(q,p) = \max\{t_d | d \in D_{p,q}\}$.

\subsection{Noise Reduction}
\label{sec:noise_reduction}

Noise reduction for BlogNEER consists of some modifications of the BaseNEER approach as well as additional filtering steps. The filters take place prior to (a-priori) and after (a-posteriori) the finding of temporal co-references, the core of the BaseNEER method. However, our experiment indicated that the NEER co-reference detection itself also causes additional noise in the result set. This is mainly caused by the way terms are merged into classes using different rules as well as by the co-occurrence consolidation among the merged classes. By changing how rules are applied and substituting co-reference classes with sub-term classes, which are more diffident in grouping terms, we achieve a better result. 

\subsubsection{BaseNEER Modifications}
\label{sec:NEER_modifications}

\begin{sloppypar}
In BaseNEER, terms are merged into co-references classes by applying four rules: 1. Prefix/suffix rule, 2. Sub-term rule, 3. Prolong rule, 4. Soft sub-term rule (for definitions s. \cite{Tahmasebi2012}, Section\ref{sec:BaseNEER}). The first three rules are applied iteratively until no further merging can be done. Subsequently, the fourth rule is applied for a final merging. Each rule results in co-reference classes consisting of the merged terms and represented by the most frequent term. After each rule, all co-reference classes with the same representative are merged into one. 
\end{sloppypar}

While the sub-term rules (rule 2 and 4) most likely find real co-references as one of the two merged terms is entirely included in the other, the prefix/suffix merging (rule 1) is rather vague. For example, \textit{Chancellor Angela Merkel} and \textit{Chancellor Gerhard Schroeder} are merged together, although they obviously do not refer to the same person. Angela Merkel is the current chancellor of Germany, however, Gerhard Schroeder is her predecessor.

\begin{sloppypar}
After applying the rules, when all co-reference classes have been created, a consolidation of the co-occurrence relations among the terms in a co-reference class is performed (s. Fig.~\ref{fig:NEER_workflow}, Finding Temporal Co-references). During this step, the representative of a class is connected to the terms co-occurring with any term in its co-reference class. This creates a graph with the co-reference classes as nodes and the edges representing co-occurrences among the terms of two co-reference classes. Connected classes are considered as potential indirect co-references of each other.
\end{sloppypar}

\begin{sloppypar}
By having false positives in a co-reference class,  the consolidation causes even more noise as the co-reference class is connected to more terms, being considered potential co-references. Using the example above, while \textit{Chancellor Angela Merkel} separately would only be connected to the terms it co-occurs with, after the merging and consolidation it is also connected to the terms co-occurring with \textit{Chancellor Gerhard Schroeder}. As both terms are not true co-references of each other, it is less likely for the co-occurrences of \textit{Chancellor Gerhard Schroeder} and \textit{Chancellor Angela Merkel} to be true co-references. Therefore, for BlogNEER, the prefix/suffix rule is discarded to prevent the described behavior.
\end{sloppypar}

In contrast to the prefix/suffix rule and the two sub-term rules, the prolong rule is not intended to find temporal co-references among the extracted terms. Instead, it creates new terms by merging two terms into a longer term. The prerequisite for the merging is that the terms have a lexical overlap and some prefix of the newly created term has a co-occurrence relation to the remaining suffix. For example, \textit{Prime Minister} and \textit{Minister Tony Blair} have a lexical overlap (i.e., \textit{Minister}). The resulting term \textit{Prime Minister Tony Blair} can be split up into \textit{Prime Minister} and \textit{Tony Blair}. As these two terms have a co-occurrence relation in the considered dataset, the prolong rule would be applied to create the new term \textit{Prime Minister Tony Blair}. All three terms are put together into a co-reference class for further processing.

Even though this rule can find new temporal co-references, like in the example above, it also creates noise. Think of the term \textit{Prime Minister Blair}, which co-occurred in our dataset with the term \textit{Blair Witch Project}. Applying the prolong rule in the same manner as in the example above, creates \textit{Prime Minister Blair Witch Project} which is obviously false. Often this phenomenon was caused by mistakes during the entity extraction process, like in the sentence ``we met Obama Friday'' where \textit{Obama Friday} was extracted and resulted in \textit{Barack Obama Friday} after applying the prolong rule. Such a term is very difficult to filter out because it does not exist and thus, knowledge bases cannot provide semantic information. Also it is considered fairly high frequent as it sums up the frequencies of the terms \textit{Prime Minister} and \textit{Tony Blair}, which are both frequent for the query \textit{Prime Minister Blair}. Therefore, a frequency filter is not useful in this case.

However, by disabling the rule completely we would lose long terms like \textit{Prime Minister Tony Blair}.\footnote{terms of length 4, like Prime Minister Tonly Blair, are intentionally not extracted during the extraction phase to reduce the amount of noise.} Therefore, instead of taking all generated rules into our created context, we check the existence of the term first. Since we incorporate DBpedia on the semantic filtering, we use it for the existence check as well. Hence, a term created by the prolong rule will only be considered for further processing if there is a corresponding resource on DBpedia.

The merged terms result in co-reference classes consisting of the direct co-references, which, when applied to Web data, contained a large amount of noise. We would end up with the terms \textit{Brown} and \textit{Prime Minister Brown} as direct co-references of \textit{Prime Minister Blair}. Even though we did not apply the prefix/suffix rule, the found direct co-references are only related by an overlap of a prefix (\textit{Prime Minister}).

\begin{sloppypar}
\begin{enumerate}
\item \underline{Sub-term rule:}\\
Prime Minister Blair [\textit{Blair}] $\Leftrightarrow$ Prime Minister [Minister]\\
$\Longrightarrow$ Prime Minister [\textit{Prime Minister Blair, Minister, Blair}]
\item \underline{Sub-term rule:}\\
Prime Minister Brown [\textit{Minister, Brown}] $\Leftrightarrow$ Prime Minister [Minister]\\
$\Longrightarrow$ Prime Minister [\textit{Prime Minister Brown, Minister, Brown}]
\item \underline{Merging classes with same representative:}\\
$\Longrightarrow$ Prime Minister [\textit{Prime Minister Blair, Minister, Prime Minister Brown, Blair, Brown}]
\end{enumerate}
\end{sloppypar}

\begin{sloppypar}
To prevent such a merging we do not consider the terms in the merged classes as direct co-references. Moreover, instead of merging terms into co-reference classes, we introduce sub-term classes. A sub-term class only includes terms that consist of sub-terms (or super-terms respectively) of each other. Instead of having only the most frequent term of a co-reference class as representative, every term represents its own sub-term class. In order to achieve this, we do not remove classes that have been merged with another class. Every sub-term class is represented by the longest term in that class, which is the super-term of all other terms in that class. Thus, with the terms from the example above we end up with eight classes:
\end{sloppypar}

\begin{sloppypar}
\begin{itemize}
\item Prime Minister Tony Blair [\textit{Prime Minister Blair, Prime Minister, Minister, Tony Blair, Tony, Blair}]
\item Prime Minister Blair [\textit{Prime Minister, Minister, Blair}]
\item Prime Minister Brown [\textit{Prime Minister, Minister, Brown}]
\item Prime Minister [\textit{Minister}]
\item Tony Blair [\textit{Blair}]
\item Brown []
\item Blair []
\item Tony []
\end{itemize}
\end{sloppypar}

\begin{sloppypar}
Eventually, the members of all sub-term classes that contain a certain term are considered direct co-references of this term. Let $super_w$ be the set of all super-terms of $w$, then $w$ is contained in the sub-term classes $sub_s$ for all $s \in super_w$. Accordingly, the direct co-references of $w$ are all terms in these sub-term classes:
\[direct\_corefs(w) = \bigcup_{s \in super_w} sub_s\]
\end{sloppypar}

With the example context around \textit{Tony Blair} we obtain the following direct co-references for the involved terms:

\begin{sloppypar}
\begin{itemize}
\item \textbf{Prime Minister Tony Blair}: Prime Minister Blair, Prime Minister, Minister, Tony Blair, Tony, Blair
\item \textbf{Prime Minister Blair}: Prime Minister Tony Blair, Prime Minister, Minister, Tony Blair, Tony, Blair
\item \textbf{Prime Minister Brown}: Prime Minister, Minister, Brown
\item \textbf{Prime Minister}: Prime Minister Tony Blair, Prime Minister Blair, Prime Minister Brown, Minister, Tony Blair, Tony, Blair, Brown
\item \textbf{Minister}: Prime Minister Tony Blair, Prime Minister Blair, Prime Minister Brown, Prime Minister, Tony Blair, Tony, Blair, Brown
\item \textbf{Tony Blair}: Prime Minister Tony Blair, Prime Minister Blair, Prime Minister, Minister, Tony, Blair
\item \textbf{Brown}: Prime Minister, Minister
\item \textbf{Blair}: Prime Minister Tony Blair, Prime Minister Blair, Prime Minister, Minister, Tony Blair, Tony
\item \textbf{Tony}: Prime Minister Tony Blair, Prime Minister Blair, Prime Minister, Minister, Tony Blair, Blair
\end{itemize}
\end{sloppypar}

\begin{sloppypar}
Although the direct co-references are computed as union of multiple classes, the consolidation of co-occurrence relations is still only performed among the terms in a sub-term class. The rationale behind this is that every co-occurrence of a term $w$ is also a co-occurrence of the sub-terms of $w$, but the inverse does not hold. Consider the following sentence:
\end{sloppypar}
``Kinect, formerly known as Project Natal, ...''

\begin{figure}[h]
\centering
\includegraphics[width=\columnwidth]{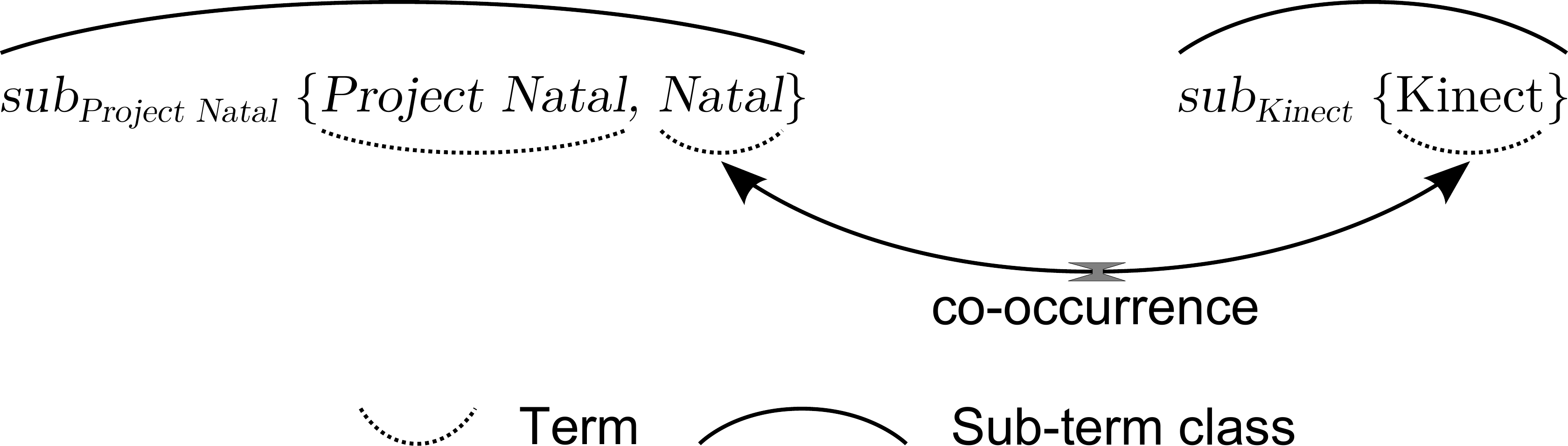}
\caption{Co-occurrence relation before consolidation.}
\label{fig:before_consolidation}
\end{figure}

As \textit{Kinect} co-occurs with \textit{Project Natal} it also co-occurs with \textit{Natal}. However, the same sentence containing just \textit{Natal} instead of \textit{Project Natal} only leads to a co-occurrence relation between \textit{Kinect} and \textit{Natal}, but not \textit{Project Natal}, as shown in Fig.~\ref{fig:before_consolidation}.

\begin{figure}[h]
\centering
\includegraphics[width=\columnwidth]{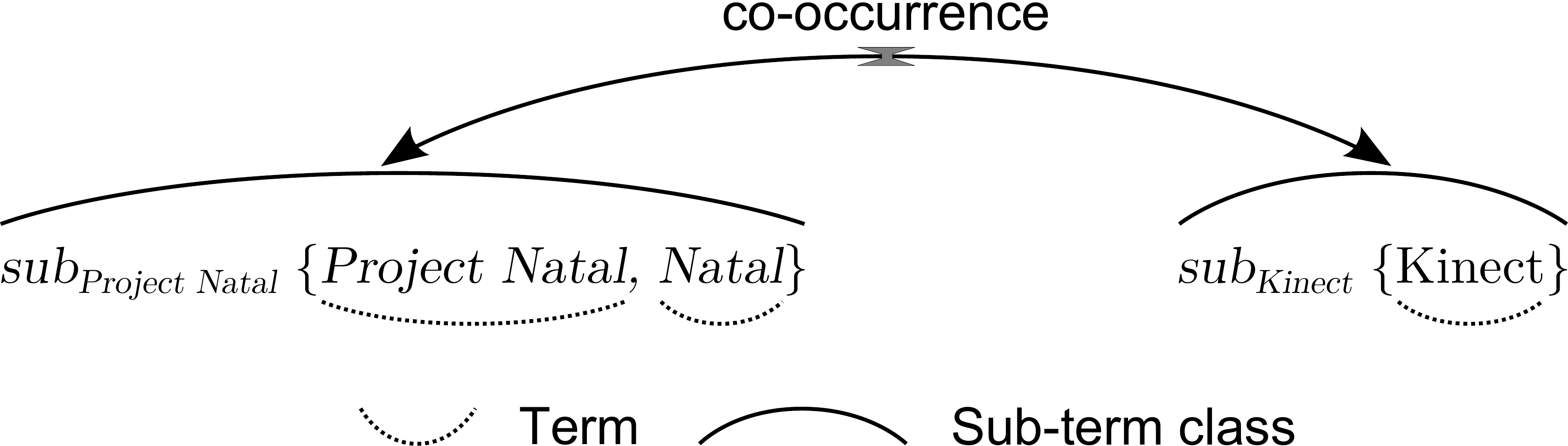}
\caption{Co-occurrence relation after consolidation.}
\label{fig:after_consolidation}
\end{figure}

To obtain this co-occurrence relation, we consolidate the relations of the terms in a sub-term class and connect the sub-term classes accordingly, as shown in Fig.~\ref{fig:after_consolidation}.

After consolidating the co-occurrence relations of all terms in a sub-term class $\mathit{sub}_w$, we consider the representatives $r_s$ of the connected sub-term classes $s \in \mathit{sub}_w$ as co-references of $w$:
\[\mathit{corefs}(w) = \{r_s\ |\ s \in \mathit{related}(\mathit{sub}_w)\}\] 

The resulting set also includes direct co-references, as every sub-term inevitably co-occurs with its super-terms. Therefore, to obtain the indirect co-references, we need to subtract the direct co-references from the set of all co-references of a term $w$:
\[\mathit{indirect\_corefs}(w) = \mathit{corefs}(w) \setminus \mathit{direct\_corefs}(w)\]

\subsubsection{Frequency Filtering}
\label{sec:frequency_filtering}

Not all co-occurring terms can be considered name evolutions of each other. However, during the co-reference detection step, as described before, we consider them all to be potential names of the same entity. Therefore, the erroneously detected candidates need to be filtered. Although the semantic filter that is applied at the very end of the BlogNEER pipeline is capable of filtering out false positives, it can only handle known terms that semantic information are available for. Hence, a pre-filtering of infrequent terms increases will lead to a better accuracy in the end. To some part, this problem is tackled by BaseNEER with a basic a-priori frequency filtering. The filter prevents BaseNEER from taking misspelled terms into account by filtering out terms with a total document frequency below a threshold. This however, is not applicable for Web data; an analysis of our Blog datasets revealed that term frequencies vary strongly. Some query terms are mentioned in many more documents and by many more sources than others. Therefore, too low parameter values would not be sufficient to filter out the amount of noise needed for achieving satisfactory precision values with frequent query terms. In contrast, by increasing the thresholds and hence adapt them to more frequent terms, we drastically lowered the recall for infrequent query terms. Therefore, we opted for a dynamic approach that automatically adjusts the parameters according to the varying number of documents retrieved for different query terms. 

\begin{figure}[t]
\centering
\includegraphics[width=\columnwidth]{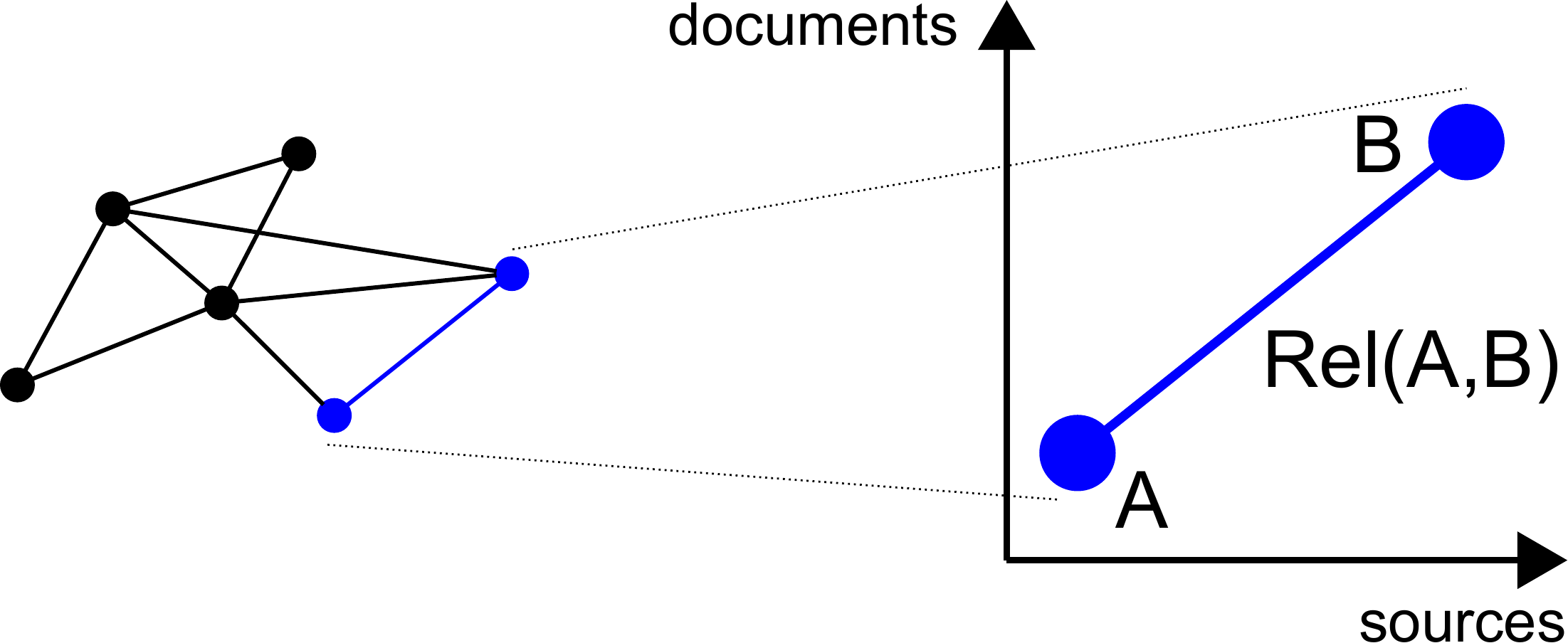}
\caption{Every node as well as every relation among two nodes in a context graph has a document and a source frequency.}
\label{fig:graph_relation}
\end{figure}

In contrast to the dataset that was used for the evaluation of BaseNEER and consisted of only one source (i.e., New York Times) the Web consists of numerous sources. Therefore, when a name of an entity changes, multiple sources are likely to report about this event. While all of them may use different words in their articles, the most common terms related to a query are used by all or most sources. Furthermore, also co-occurrence relations among two terms feature these characteristics, i.e., to appear in different documents and sources. As described in Sect.~\ref{sec:BaseNEER}, during the NEER process, a graph is build consisting of extracted terms where all pairs of terms that co-occur within a certain distance are connected. Fig.~\ref{fig:graph_relation} shows such a graph and one of its edges in a diagram indicating the document and source frequencies for the two terms.

With this in mind we introduce the following dynamic thresholds as parameters for the frequency filtering:

\begin{itemize}
\item \textbf{$\mathit{minDocFr}$}: minimum total document frequency of a term across all sources.
\item \textbf{$\mathit{minSrcFr}$}: minimum source frequency of a term.
\item \textbf{$\mathit{minRelSrcFr}$}: minimum source frequency threshold of a co-occurrence relation.
\end{itemize}

\begin{sloppypar}
As basis for the dynamic thresholds adjustment we use the total number of sources for which documents are retrieved as well as the lowest document frequency among the query's sub-terms. Another parameter specifies the number of terms we want to receive. For instance, for the query term \textit{President Obama} we regard the number of sources that mention the query term and the number of documents containing the sub-term (i.e., \textit{President} and \textit{Obama}) that was least mentioned during the specified change periods. The actual parameters are set to a fraction of these values and lowered stepwise until we have fetched the number of terms we aim to build up the context with.
\end{sloppypar}

At each step during the adjustment we pick the relevant terms and relations. For terms relevant  means being related to the query $q$, i.e., a term $w$ occurs in a sufficient number of documents or sources with respect to the parameters and $q$. Additionally, a term is considered relevant if it includes a sub-term of the query. For relations, relevant means that the terms by the relation are considered relevant to the query or the relation itself is frequent with respect to the sources it occurs in. As we derive co-references based on co-occurring terms, we are interested in the relevant terms as well as their co-occurrence relations. All terms or relations respectively that do not meet one of these relevance conditions are filtered out. The remaining co-occurrence relations constitute the context graph that is passed on to the co-reference detection step. 

Another frequency filtering takes place after the co-reference detection. This a-posteriori filter is based on the consolidated frequencies of the derived sub-term classes. The frequency of a sub-term class is the sum among all terms in that class. Also the weights of the edges in the context graph, representing the co-occurrence relations among terms of two classes, are summed up when they are consolidated (as described in Sect.~\ref{sec:NEER_modifications}). By that new frequencies are provided to filter on. The maximum of these values are scaled and used as lower bounds. All co-references of which its sub-term class falls below these lower bounds are filtered out.

The a-posteriori frequency filter is not applied to the full context graph. Since the graph has reached its final state in the previous step of the NEER process (s. \ref{fig:filtering_hooks}, Co-reference Detection), we only consider the sub-graph around the query term $q$. This consists of the sub-term class of $q$ and all connected sub-term classes, represented by the co-references of $q$. The co-references can be regarded as an unfiltered set $\mathit{unfiltered}_q$ of 4-tuples. Every tuple consists of the co-reference term $w \in \mathit{corefs}(q)$, the aggregated document frequency $\mathit{df}_w$ of $\mathit{sub}_w$, the consolidated source frequency $\mathit{sf}_w$ of the relation between $\mathit{sub}_q$ and $\mathit{sub}_w$ as well as the sub-terms of $w$:
\[\mathit{unfiltered}_q = \{(w,\ \mathit{df}_w,\ \mathit{sf}_w,\ \mathit{sub_w})\ |\ w \in \mathit{corefs}(q)\}\]

\begin{sloppypar}
As foundation for the filtering, we determine the maximum frequency $\mathit{df}_w$ and maximum relation source frequency $\mathit{sf}_w$ among all co-reference terms $w \in \mathit{corefs}(q)$:
\[\mathit{max\_df}_q = \max_{w \in \mathit{corefs}(q)} \mathit{df}_w\] 
\[\mathit{max\_sf}_q = \max_{w \in \mathit{corefs}(q)} \mathit{sf}_w\]
\end{sloppypar}

These two values multiplied with parameters $k$ and $l$ yield lower bounds for the document and source frequencies of co-references to be considered in the further process. Accordingly, we filter out co-references $w$ with a document frequency $\mathit{df}_w < \mathit{max\_df}_q \cdot k$ and a source frequency $\mathit{sf}_w < \mathit{max\_sf}_q \cdot l$. As the source frequency $\mathit{max\_sf}$ represents the weight of an edge from $q$ to one of its co-references, $l$ is denoted $\mathit{weight\_factor}$. As $k$ determines the frequencies of the filtered co-references, it is denoted $\mathit{frequency\_factor}$. Both can have values between 0 and 1. For instance, a $\mathit{weight\_factor}$ of 0.5 means, we only keep co-references that co-occur in at least half as many sources as the most common co-reference with respect to sources it co-occurs in. $\mathit{frequency\_factor}$ of 0.5 on the other hand means we keep all co-references that are mentioned in at least half as many documents as the most common co-reference with respect to the documents it co-occurs in.

The remaining set of filtered co-references $\mathit{filtered}_q$ for the query term $q$ is considered the preliminary final result, to be filtered by further filters (i.e., Semantic Filtering, s. Section~\ref{sec:semantic_filtering}):
\begin{multline*}
\mathit{filtered}_q = \{\mathit{coref}_w \in \mathit{unfiltered}_q|\\
\mathit{df}_w \leq \mathit{max\_df} \cdot k \lor \mathit{sf}_w \leq \mathit{max\_sf} \cdot l\}
\end{multline*}

\subsection{Semantic Filtering}
\label{sec:semantic_filtering}

\begin{sloppypar}
Semantic Filtering is a novel filtering method for NEER incorporating the Semantic Web. We use semantic information from DBpedia to augment a term to identify names that do not refer to the same entity. The rational behind this filter is the following; two terms referring to entities of different types or categories can not be evolutions of each other. By taking this information into account, all co-references identified as names for entity types different than the query term will be filtered out. In addition to types and categories we consider any year from the semantic property values of a term and compare it with the years of the query term. In case of two names referring to the same entity they must have years in common (e.g., birth date) or in case of name evolution the end year of the old name must match the year of the new name's introduction.
\end{sloppypar}

Among the available knowledge bases, e.g., DBpedia, Yago, Freebase, DBpedia fits our needs best. It facilitates the use for our purpose by integrating semantic information about names in its architecture, such as redirections from alternative names of an entity. Freebase is based on IDs instead of names and therefore there is no trivial mapping from an entity's name to a resource. In order to find all entities corresponding to a name, a string search on the name and alias properties of an entity would be required. These properties often only contain the full names and are therefore a limitation for NEER. For example, the former pope \textit{Pope Benedict XVI} is often mentioned in articles as \textit{Pope Benedict} because the \textit{XVI} is implicitly understood. However, \textit{Pope Benedict} is not a real alias and therefore not available on Freebase. On the other hand, a partial string match would yield many more entities and therefore also is not suitable. Yago on the other hand, provides names to access entities but uses only one unique name as identifier for a resource. For instance, \textit{Joseph Ratzinger}, the original name of \textit{Pope Benedict} does not match any resource because its identifier is, again, \textit{Pope Benedict XVI}. For BlogNEER, we depend on a certain level of semantic information about names, such as redirections from alternative names and possible disambiguations candidates for ambiguous names like \textit{Pope Benedict} (without its identifying number). DBpedia provides these properties with disambiguation pages and redirections from alternative names to a resource. In addition, the information on DBpedia is derived from Wikipedia and created by its user community and is therefore seen as a reliable source. As these elements are essential for our approach, we make use of DBpedia for BlogNEER. 

On DBpedia every resource has its own unique name and every name only directly points to one resource. For ambiguous terms there exist separate disambiguation resources, e.g., \textit{Apple\_(disambiguation)} is the disambiguation resource of the term \textit{Apple}. This disambiguation resource contains links to the resources \textit{Apple} (the fruit) and \textit{Apple\_Inc}. Unlike this example, disambiguation resources do not always have the ``disambiguation'' suffix. However, every resource has properties, which either point to a textual or numeric value, or to another resource. Disambiguation resources can be identified by the existence of disambiguation properties that point to their corresponding unambiguous resource.

Other properties which are important for our work is the \textit{type} as well as the \textit{subject}  of a resource, which can be conceived as categories. In addition to the properties (resource $\rightarrow$ property $\rightarrow$ value), DBpedia also provides the inverse properties (value $\rightarrow$ is-property-of $\rightarrow$ resource).

By mapping a query term as well as all of its co-references (direct and indirect) to DBpedia resources, terms can be augmented with their semantic properties. These properties can help to filter out false positive results derived during the NEER process. Here it is important to note that only descriptive properties will be used and known name evolution information and co-references information from DBpedia will not be utilized. Although, in some cases redirects represent a name change as well by redirecting an old name to its new name, we do not use this information explicitly. Hence, we treat all terms separately, even if they redirect to the same resource, like there is no redirection available (e.g., for \textit{Czechoslovakia} and \textit{Czech Republic} or \textit{Slovakia}).

\subsubsection{Resolving Names}
\label{sec:resolving_of_names}

The process of mapping a term to its corresponding semantic resource on DBpedia is called \textit{name resolving}. In case a term cannot be found on DBpedia, we incorporate the direct co-references to resolve the term. Even though this does not work in all cases, the heuristic provides good results.

\begin{sloppypar}
For resolving a name, we order its direct co-references by descending length and by ascending frequency. Thus, we try to resolve longer terms first. If two terms are of same length, we take the less frequent one first. This is intended because longer terms as well as less frequent terms are more expressive. For example, consider the query \textit{Mr John Doe}. The term cannot be resolved, as it does not exist on DBpedia. Therefore, we try to resolve its direct co-references \textit{John Doe}, \textit{John} and \textit{Doe}. Most likely, \textit{John} is the most frequent term as it is a very common name. Therefore, the semantic resource of \textit{John} probably does not represent \textit{Mr John Doe}. Since we take the longest term first, we consider \textit{John Doe} next. This actually exists and refers to the right entity for our query. In case where \textit{John Doe} not exists either, we would have to decide between \textit{John} and \textit{Doe}. Since \textit{John} is more frequent, we would try to resolve \textit{Doe} next. This is again more expressive, as there exist less persons with name \textit{Doe} than with the name \textit{John}. However, as \textit{Doe} is still an ambiguous name we will need to disambiguate it to \textit{John Doe} after all, as described in the next subsection.
\end{sloppypar}

\subsubsection{Disambiguation and Aggregation of Properties}
\label{sec:disambiguation}

\begin{sloppypar}
Once we resolved a name successfully, we fetch all properties as well as the inverse properties and save them in a lookup table. In this table, every property gets indexed twice; once by the complete property URI (e.g., http://www.w3.org/1999/02/22-rdf-syntax-ns\#type, short \textit{rdf:type}) and once by the name extracted from the URI (e.g., type). Both keys point to the value of the property. By indexing the properties names in addition to the unique identifiers we are able to retrieve a list of all types independently from the used ontology. This is important since some resource have assigned the same properties from different ontologies (e.g., http://dbpedia.org/property/type in addition to \textit{rdf:type}). By using the name (e.g., \textit{type}) for indexing, we unify the properties from different ontologies to the one property.
\end{sloppypar}

After mapping the found terms to their corresponding resources, we follow four strategies to extend and disambiguate what entities they refer to. The first strategy is to follow DBpedia redirections if any redirections are present. The second strategy is to explore disambiguation resources for ambiguous terms that do not redirect to a disambiguation resource directly (e.g., \textit{Apple} from the example above). The remaining two strategies disambiguate ambiguous terms.

\begin{figure}[h]
\includegraphics[height=1.35cm]{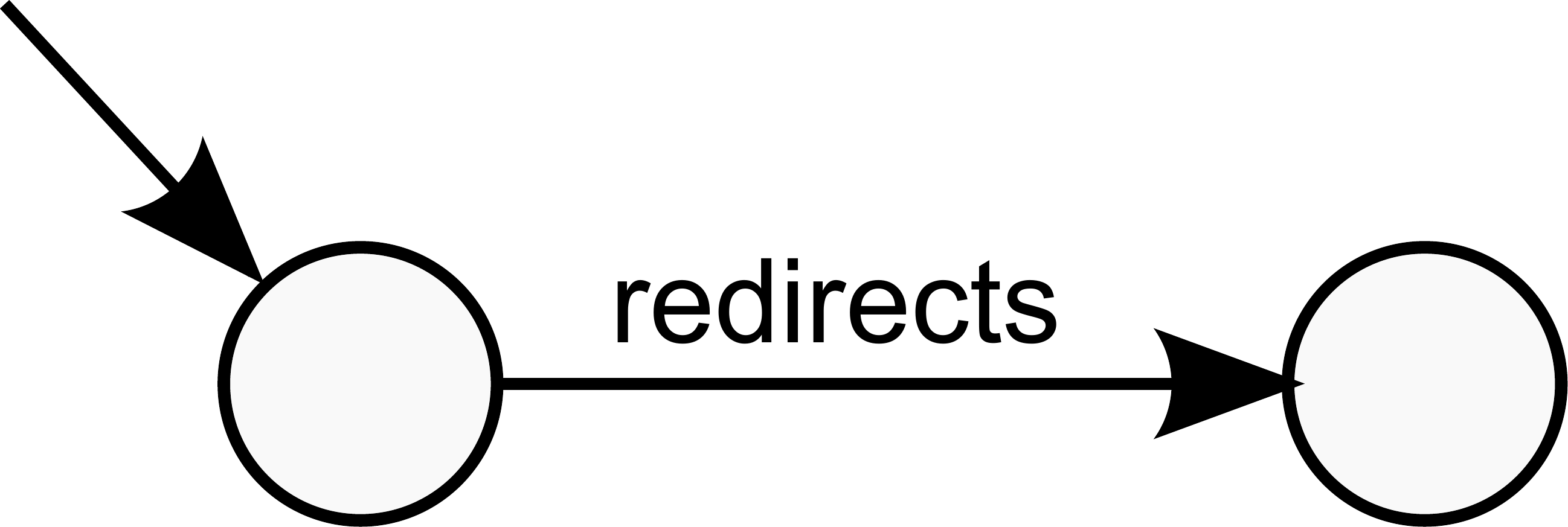}
\caption{Following redirections.}
\label{fig:DBpedia_redirects}
\end{figure}

\begin{sloppypar}
{\bf Redirection Strategy} Redirections are realized on DBpedia by a redirection property (i.e., http://\allowbreak{}DBpedia.org/\allowbreak{}ontology/\allowbreak{}wiki\allowbreak{}Page\allowbreak{}Redirects, short \textit{dbpedia-owl:wikiPageRedirects}). This is assigned to the resource that is supposed to redirect to another. We leverage this by fetching the resource that the property points to (s. Fig.~\ref{fig:DBpedia_redirects}). Redirects are followed recursively. During this procedure we fetch and index all new found properties and aggregate them. The rationale behind this is that, in case there is a redirection pointing to another resource, this resource is supposed to give a better description of the corresponding entity. Therefore, it represents the same entity and its properties belong to this entity as well.
\end{sloppypar}

{\bf Ambiguation Strategy} If a resource has an ambiguous meaning, it mostly redirects to a disambiguation resource. In this case, we apply the first redirection strategy. However, there are ambiguous resources that do not redirect. For instance, the resource \textit{Apple} represents the fruit, even though \textit{Apple} is an ambiguous term. The disambiguation resource for \textit{Apple} is \textit{Apple\_(disambiguation)}, but there is no redirection between these two. However, as \textit{Apple\_(disambiguation)} uses the \textit{dbpedia-owl:wikiPageDisambiguates} property to point to its non-ambiguous resources, like \textit{Apple} (the fruit), there is an inverse property on the resource of the fruit.

\begin{figure}[h]
\includegraphics[height=3cm]{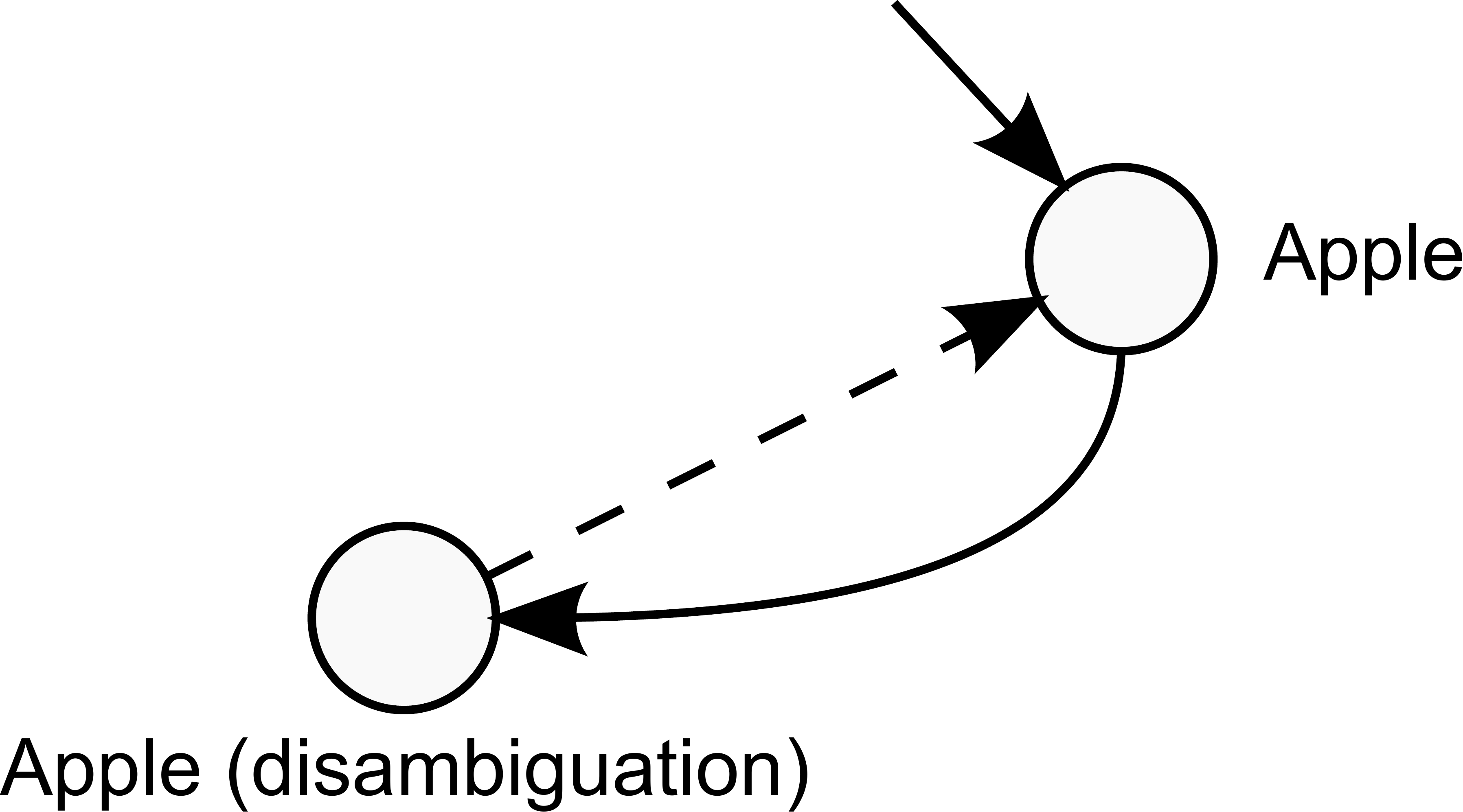}
\caption{Redirecting to a disambiguation resource.}
\label{fig:DBpedia_ambiguates}
\end{figure}

To discover ambiguous terms, we analyze all inverse disambiguation
relations of a resources and follow backwards if there is a relation
originating in a resource with the exact same name as the original
term, but with the suffix ``(disambiguation)'' appended (s. Fig.~\ref{fig:DBpedia_ambiguates}). Unlike for the redirection, we do not collect all properties. Instead, we only keep the properties of the disambiguation resource, because the original term might not the one we are interested in (e.g., Apple fruit).

\begin{figure}[h]
\includegraphics[height=3cm]{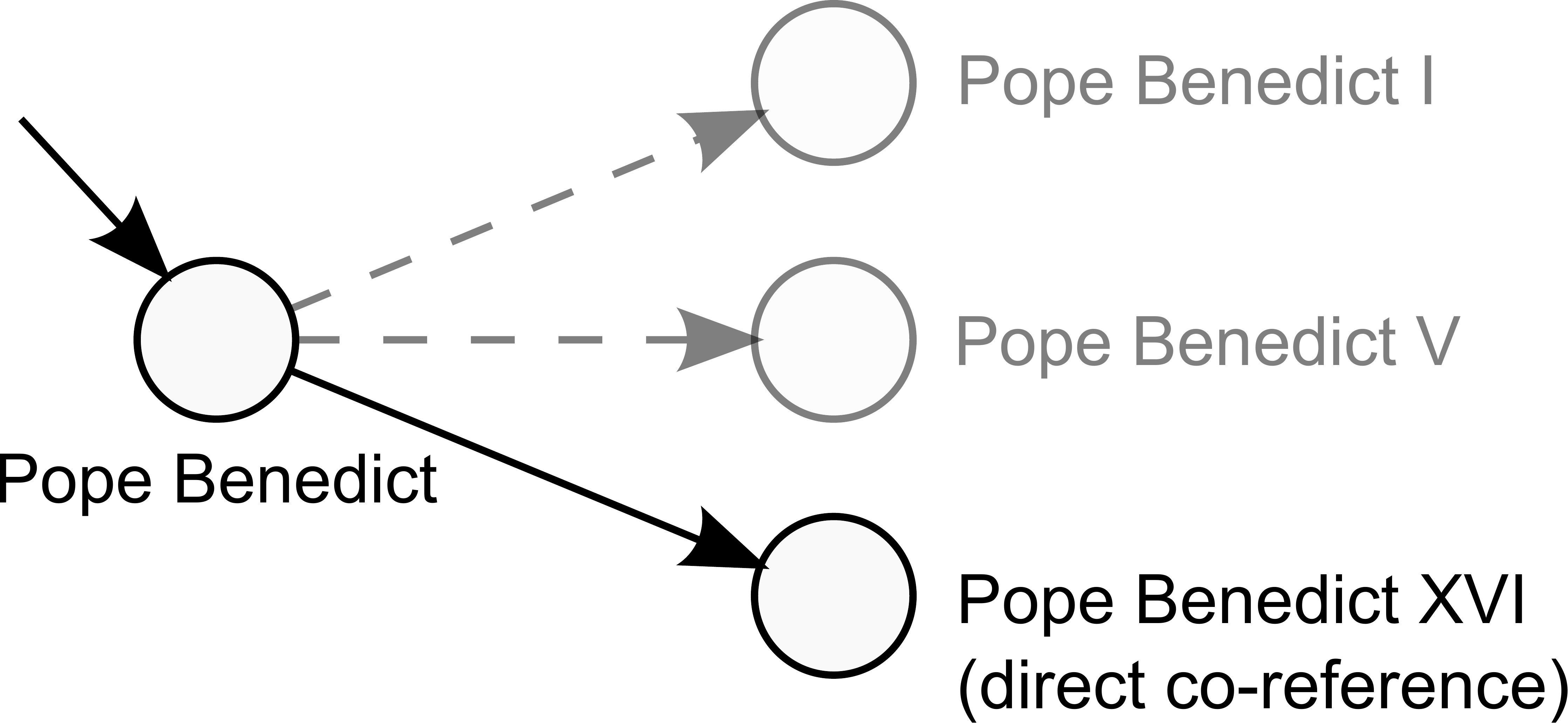}
\caption{Disambiguation by incorporating direct co-references.}
\label{fig:DBpedia_direct_disambiguation}
\end{figure}

\begin{sloppypar}
{\bf Direct Disambiguation Strategy} If a disambiguation resource has been identified, we need to decide for one of the suggested resources as representation of the entity under consideration. In case one of the candidates proposed by DBpedia is also a direct co-reference of the entity or its name respectively, we take this one as representing resource. This is illustrated by the example in Fig.~\ref{fig:DBpedia_direct_disambiguation}. The term we try to resolve in this example is \textit{Pope Benedict}. The corresponding disambiguation resource proposes all popes with name Benedict I up to Benedict XVI. As \textit{Pope Benedict XVI} is a direct co-reference of \textit{Pope Benedict} we follow this resource, like we do in our redirection strategy by aggregating its properties with the properties that have been fetched so far.
\end{sloppypar}

\begin{sloppypar}
\end{sloppypar}

\begin{figure}[h]
\includegraphics[width=\columnwidth]{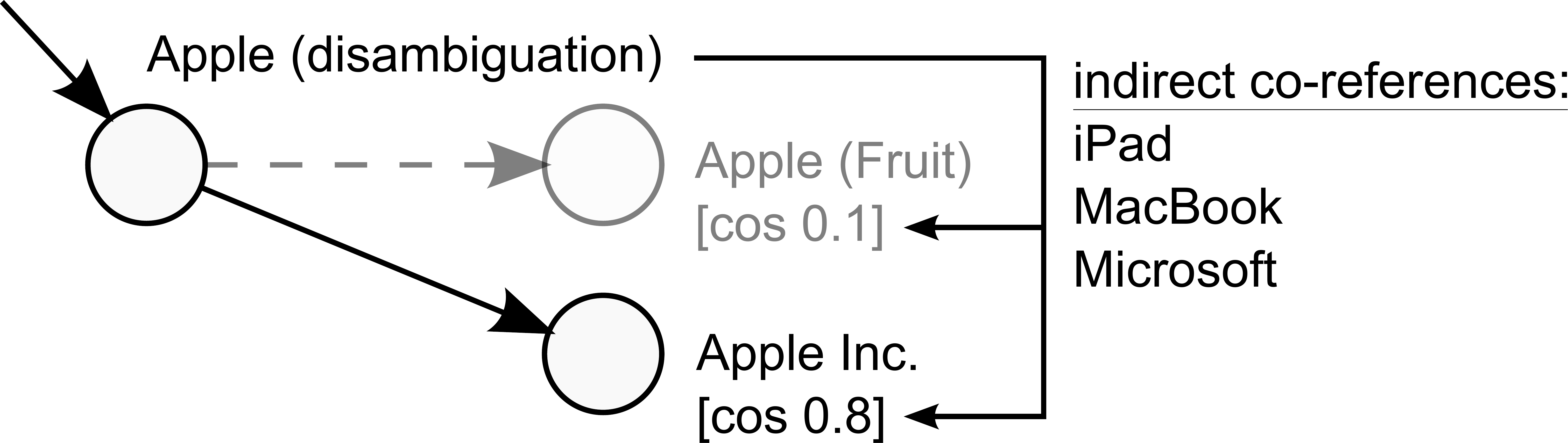}
\caption{Disambiguation by following the most similar resource.}
\label{fig:DBpedia_indirect_disambiguation}
\end{figure}

\begin{sloppypar}
{\bf Indirect Disambiguation Strategy}
In case none of the detected direct co-references is listed as disambiguation candidate on DBpedia, the ambiguous term is disambiguated based on its similarity to the suggested candidates. Therefore, we make use of the indirect co-references $ind_1$, $ind_2$, \dots to form a term vector. Every dimension of this vector $v_{ind}$ represents one indirect co-reference and is weighed with its occurrence frequency as obtained during the co-reference detection phase: $(freq(ind_1),\ freq(ind_2),\ \dots)$. Additional vectors $v_{{candidate}_i}$ are created for each disambiguation candidate with the same dimensions as before. The weights of these vectors represent the occurrences of the indirect co-references in values of the candidate's semantic properties: $(occur(ind_1, i),\ occur(ind_2, i),\ \dots)$. Similar to \cite{Garca2009} we calculate the cosine similarity between each candidate vector and the vector of indirect co-references to measure which resource fits the ambiguous term best:
\[cos_i = \dfrac{v_{ind} \cdot v_{{candidate}_i}}{|v_{ind}| \cdot |v_{{candidate}_i}|}\]
The resource of the most similar candidate will be selected as the semantic representation for the term. This procedure is illustrated in Fig.~\ref{fig:DBpedia_indirect_disambiguation}.
\end{sloppypar}

\subsubsection{Filtering}
\label{sec:filtering}

After the disambiguation and aggregation of properties from DBpedia, we proceed with the filtering. For this, we consider the properties \textit{type} and \textit{subject} regardless of their ontology, as described in Sect.~\ref{sec:disambiguation}. For the filtering, we treat DBpedia under the \textit{open world assumption}. That means, the fact that a resource does not have a certain property does not mean its corresponding entity does not have the property either. Perhaps the resource has just not been annotated with the property. However, if a resource has a certain property, we consider this to be complete. For instance, if a resource is annotated with types, we assume these are all types it has and there is no type missing.

{\bf Similarity Filtering} The first filter we apply to the set of candidates considers the similarity between the query term and its co-reference candidates based on the their types, subjects and the years they have in common. We compare the set of types and subjects of the query term with the same sets of each co-reference candidate. This methodology is limited in that it requires query terms, or the corresponding DBpedia resource respectively, and their detected co-references share the same properties. Otherwise, the semantic filtering method is not applicable. It would be wrong to consider two terms referring to different entities just because one of them has not been annotated with types or subjects while the other one has (open world assumption, s. above). In case both the query term as well as the co-reference under consideration have been annotated with types or subjects, we require them to have at least one type and/or subject in common. As an alternative for subjects we use the years derived from the properties of a resource. If both resource that we compare have been annotated with years as property values, we compare these instead of subjects. The rationale is that an entity can change its subject when changing its name, for example persons when they get a new office. Nevertheless, they will have years in common, like the end and start date of the new and old office. Additionally, even if two entities have subjects in common, they may be distinguished by different years. All co-reference that have not been filtered out are passed to a type hierarchy filter.

\begin{sloppypar}
{\bf Type Hierarchy Filtering} Different from the similarity filter, the type filter considers hierarchies of types in addition to the types a resource is directly annotated with. For instance, as both \textit{Pope Benedict XVI} and \textit{Barack Obama} are persons, the similarity filter would not have filtered out one of them as co-reference of the other. However, \textit{Pope Benedict XVI} is of type \textit{dbpedia-owl:Cleric} while \textit{Barack Obama} is annotated with \textit{dbpedia-owl:OfficeHolder}. Both types are sub-types of \textit{Person}. Thus, the two terms refer to different kinds of persons on DBpedia and do most likely not correspond to the same entity.

For this filtering we analyze the sub-class relations of all types assigned to a resource. Each type on DBpedia is represented as an URI that points to a resource of that type. To obtain the hierarchy of a type, we leverage the \textit{rdfs:subClassOf} property of this resource. This points to its super-type and allows us to perform the procedure recursively until there is no further \textit{rdfs:subClassOf} property available or no resource for the type URI exists.
\end{sloppypar} 

After having fetched the hierarchies for all types top-down, starting by a type and fetching the super-types, we analyze them bottom-up. For all types that the query term and its co-reference candidates have in common we compare all of their sub-types. For instance, for \textit{Pope Benedict XVI} and \textit{Barack Obama}, having type \textit{Person} in common, we compare their sub-types of type \textit{Person}: \textit{Cleric} and \textit{OfficeHolder}. As these are different we consider the two terms to not to refer to the same entity and remove them in the result set of the other. In case they are equivalent we proceed with the next sub-type. This will be done recursively as long as both terms have no sub-types in common or until they are not annotated with further sub-types.

The open world assumption holds also if the terms under consideration have a type in common but only one of them has been annotated with a further sub-types. As we cannot tell whether the sub-type is missing on DBpedia or the entity is  not actually an instance of that type, we do not filter out that co-reference and keep it in the final result set.

\section{Evaluation}
\label{sec:experiments}

In this section we will present evaluation results of BlogNEER and compare it to the results of BaseNEER. Our datasets consist of two blog collections. To obtain results that are comparable to BaseNEER, the evaluation was performed with the same test set. Both, dataset and test set are described in Sect.~\ref{sec:dataset}. As baseline for our experiments we used the evaluation results of BaseNEER applied to a newspaper dataset (i.e, New York Times) from \citet{Tahmasebi2012}.

\subsection{Dataset and Test Set}
\label{sec:dataset}

{\bf Dataset} For the evaluation we used two different blog dataset. We created one dataset ourselves by parsing blogs from Google Reader. It consists of the top 100 blogs from nine categories based on the ranking of Technorati \cite{technorati} where the categories are sports, autos, science, business, politics, entertainment, technology, living, green. The blogs were parsed for a time range from 2005 to 2013, however, not all blogs published articles during the entire frame. As we automatically parsed the blogs, we could only parse those that included the location of their news feed in the HTML code of their website. Blogs that did not provide this information were omitted. Eventually, we ended up with 8.952.855 documents from 801 blogs. This dataset is referred to as \textit{Technorati}.

\begin{sloppypar}
Additionally, we used the TREC dataset described by \citet{blogs08}. This dataset is referred to as \textit{Blogs08}. From Blogs08 we extracted English texts from the first 10\% of the feeds in the dataset. We limited on English so that the NLP tools used could be the same for BlogNEER and BaseNEER. To detect whether a text is written in English or not we scanned the first 1000 words for English stopwords, using a stopword list. Texts with more than 30\% stopwords in the scanned excerpt are considered English texts.
\end{sloppypar}

{\bf Test set} For the performance evaluation we compared our results to the extended test set that was originally created for BaseNEER \cite{testset}. It contains direct as well as indirect co-references of three categories: People, Locations and Companies. For the BaseNEER evaluation, the change periods for all the relevant name changes were identified and verified by three judges. Those that were accepted by least two judges were kept. In case an entity was annotated with a name change in January, also the previous year was added.

In order to adapt the test set to our datasets we created two separate instances of the test set, one for each dataset. In both instances we removed the query terms that do not appear in the corresponding dataset at all or that have been annotated with a change period that lies outside the time frame of the dataset. We also removed all of the expected co-references of a query term that do not appear in the corresponding dataset. Query terms with no remaining co-references were removed as well.

\subsection{Baseline Definition}

To the best of our knowledge there has not been work on Named Entity Evolution Recognition (NEER) besides BaseNEER. For this reason, BaseNEER is the only baseline we can compare to. Similar, however not comparable to NEER, is the task of Entity Linking (EL). EL tackles the problem of detecting terms and phrases in texts and linking them to knowledge bases, such as DBpedia. There has been extensive work in this direction. A recent and comprehensive survey of EL systems can be found in \citet{Cornolti2013}, who presented a framework for benchmarking the common tools in this field. However, the objective of EL is different than NEER's and a good performance based on this evaluation does not mean a good performance for NEER. While EL tools link terms to the knowledge base resource that represents and explains them best, we on the other hand want to identify common names that refer to same entities. There is a difference, caused by names as well as concepts, which often have their own semantic resources. These describe the term in the text better than the entity behind this concept or name. For instance, the best match for ``President Obama'', according to DBpedia spotlight \cite{Mendes2011}, is the DBpedia resource for ``Presidency of Barack Obama''. Therefore, considering terms and phrases that are being linked to the same entity as a result of NEER would miss out ``President Obama'' as alternative name for Barack Obama. For this reason, EL systems cannot be considered as a baseline for the particular task of NEER. However, we use similar methods as used in EL for the semantic filtering, described in Section~\ref{sec:semantic_filtering}. In this respect, EL can be considered as a foundation for BlogNEER (see. Section~\ref{sec:foundations}) rather than a competing task.

As argued above, BaseNEER is the only available baseline. However, it is not applicable to blog data due to the noisy character, which was the motivation for this work. Our attempts to run BaseNEER on either Blogs08 or Technorati both resulted in a very low precision of below 2\% for some entities, while it completely crashed for others due to too many co-occurring words and insufficient memory. Thus, we were not able to compute overall precision and recall measures for either dataset for comparison with BlogNEER. As the motivation and goal of BlogNEER was to adapt BaseNEER to Blogs in a way that it achieves similar results on blogs as BaseNEER on high quality newswire texts, we decided to compare these results for evaluating the performance of BlogNEER. Accordingly, the results of BaseNEER applied to the New York Times Annotated Corpus constitute our baseline. The comparison is based on the performance measures precision and recall, as defined in Sect.~\ref{sec:problem}.

In order to obtain values comparable to BaseNEER we adapted the way the measures are computed for BaseNEER to our data. Other than for BaseNEER, we do not reduce our test set by keeping only frequently occurring terms, because the frequencies of different entities vary strongly in our blog datasets. The BaseNEER test set only contains terms that occur at least five times in at least one change period. Thus, only these terms were considered for computing the recall. For BlogNEER, we calculate the recall dynamically based on the frequencies of the expected co-references and only consider those co-references with a frequency above a threshold. If the most frequent co-reference that we expect for an entity occurs in less than 100 documents, we use a threshold of five, just like BaseNEER. If the most frequent co-reference occurs 100 times or more though, we consider the other co-references to occur at least ten times. For entities where the most frequent expected co-reference occurs 500 times or more, we increase the threshold to 50 and for more than 1000 times, we use a threshold of 100. If an entity is mentioned extremely rarely and none of the expected co-references has a frequency above 5 but can still be found, we consider this as a full recall.

\begin{sloppypar}
Since direct co-references are slightly differently defined for BlogNEER compared to BaseNEER, we dropped the requirement of BaseNEER to find all direct but only one indirect co-reference per change period. The rationale behind this requirement was, that each indirect co-reference represents one change period. However, this does not hold for entities from the test set with only direct co-references. Therefore, we found it reasonable to treat all co-references the same. This might lead to a lower recall for BlogNEER, because we only achieve a full recall if we detect all direct as well as indirect co-references, regardless the number of change periods, but it simplifies the calculation for this and future work as it is not to be determined whether a expected co-reference in the test set is direct or indirect.
\end{sloppypar}

When computing the precision of a term, in addition to the terms from the test set, we also consider sub-terms of an expected co-reference as correct. For this, we also consider the query term itself as an expected co-reference. Otherwise, sub-terms would lower the precision, although they are inevitable in the unfiltered result set since we extract all sub-terms of a noun. For instance, \textit{Sean} is considered as a correct co-reference of \textit{Sean Combs} when we compute the precision. However, in order to achieve a full recall we require the full terms from the test set to be contained in the result.

\subsection{Experimental Setting}
\label{sec:experimental_setting}

For our experiments we created a Ruby implementation of BaseNEER, built on the foundation of the Ruby on Rails framework \cite{ruby_on_rails}, and modified it according to the BlogNEER extensions described in Sect.~\ref{sec:approach}. This developed framework provides a number of tools needed for NEER as well as an infrastructure to run experiments. Also a variety of basic Natural Language Processing functions, like word extraction and stemming, are implemented and simplified interfaces to established tools, like Stanford Named Entity Recognizer \cite{StanfordNER}, are included. In order to run experiments the framework provides a reusable job class as well as caching and logging methods.

For our experiments we ran mainly three jobs in a pipeline:

\begin{enumerate}
\item Parse Blogs
\item Detect Co-references
\item Filter Results
\end{enumerate}

\begin{figure*}[t]
\centering
\includegraphics[width=\textwidth]{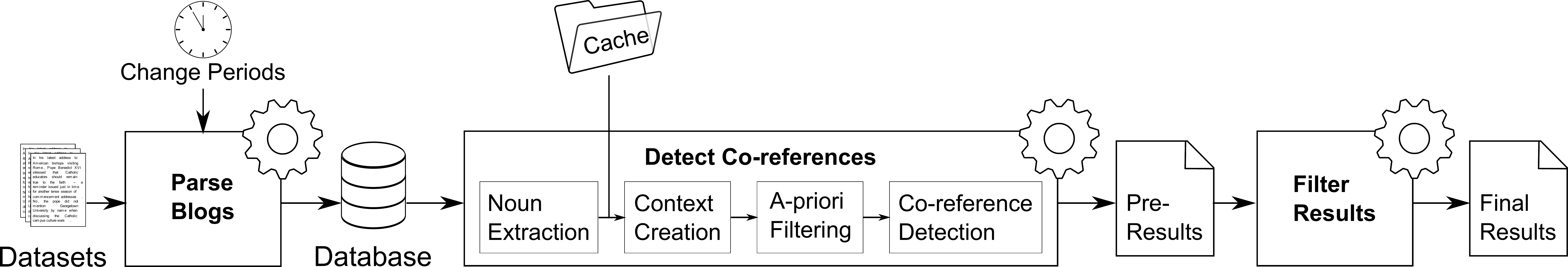}
\caption{Evaluation job pipeline.}
\label{fig:job_pipeline}
\end{figure*}

\begin{sloppypar}
The first job parsed the blog data from news feeds provided by either the Blogs08 or the Technorati dataset. The parsed data was saved to a database. Afterwards the second job accessed the database and selected the relevant data. This job was also in charge of the noun extraction, context creation and a-priori filtering. Finally, it detected the co-reference candidates by applying the NEER rules. The extracted nouns were cached to prevent another extraction of the same entities on the next run as this task is very time consuming. Lastly, the result filtering job performed the a-posteriori frequency as well as semantic filtering. The entire pipeline is shown in Fig.~\ref{fig:job_pipeline}, with the jobs denoted by the gears.
\end{sloppypar}

The NEER workflow was applied to both of our datasets and the query terms provided by the test set. Instead of detecting the change periods for the query terms in our datasets, we performed the evaluation with the same change periods that were used for the BaseNEER evaluation. These change periods are provided with the BaseNEER test set as years for the changes of each query term. The experiments with BaseNEER were performed by \citet{Tahmasebi2012} using two settings: \textit{known periods} as well as \textit{found periods}. The found periods were detected using the burst detection algorithm by \citet{Kleinberg2003}. The years of the found bursts were treated as change periods. However, these years are not included with the test set. Therefore, we only evaluated BlogNEER using the known periods and compared it to the corresponding results of BaseNEER as our baseline.

The BlogNEER algorithm and filters rely on several parameters that need to be adjusted and have an effect on the result. The first parameterizable step is the term extraction and context creation as part of the co-reference detection job. We extracted all nouns and noun phrases respectively with a word length up to three words and a minimum length of three characters. Since we are only interested in entity names, we required the terms to meet three requirements: 1. start with a capital letter, 2. do not begin or end with a stop word, 3. all words in between were required either to be stop words or to start with a capital letter. Thus, we accepted for instance \textit{Union of Myanmar}. In order to unify different spellings we ignored the case when comparing terms and stemmed the terms using a Porter Stemmer \cite{Porter1997}. To be considered as co-occurring, we required the terms to occur in a distance of at most ten words.

During the co-reference detection, which is applied on the created graph, multiple noise filters take place, as described in Sect.~\ref{sec:approach}. Some of these are static while others adjust themselves dynamically based on the extracted terms. For the dynamic filters, start values needed to be specified. The parameters determine what co-references from the context graph are kept in the final result or passed to further filtering respectively. They are crucial for the precision and recall as well as for the effort and accuracy of the semantic filtering. In the best case, all correct co-references in the result are kept while all false positives are filtered out. Having more false positives in the end result means that the semantic filter has to compare more terms, which leads to more effort and is very time consuming (with a linear growth). Moreover, this means more false positives that might not be filtered out by the semantic filter and thus make their way to the final result. To determine the parameters for the a-posteriori frequency filter we conducted an experiment that ran the filter with all combinations of values for the parameter. Afterwards the resulting co-references for each combination were filtered using semantic filtering. At the end we calculated precision and recall and picked the values that delivered the best trade-off. The exact setup of this experiment as well as all parameters we used are described in \citet{Holzmann2013a}.

\subsection{Results}
\label{sec:results}

\begin{sloppypar}
After all parameters had been determined, we ran BlogNEER on both the Blogs08 and Technorati datasets, using the described experimental setup. Our experiments resulted in a precision of 67\% and 70\% on Blogs08 and Technorati, with a fair recall of 36\% and 67\% respectively. That means, we found 36\% and 67\% of the expected co-references for the query terms from the test set, however, 67\% of what was found on Blogs08 and 70\% of what was found on Technorati is correct, which we consider the more important value.

As a baseline for our experiments we used BaseNEER applied to the New York Times dataset, as shown in Table~\ref{tab:results_baseneer}. These results are filtered using a correlation filtering and a document frequency filtering with static, partially defined parameters. Both filters are applied independently. As the Machine Learning filter approach proposed by \citet{Tahmasebi2012} must be trained and can therefore not be considered completely unsupervised, we do not compare it to our results. Table~\ref{tab:results_blogs08} and Table~\ref{tab:results_technorati} show the results of BlogNEER applied to Blogs08 and Technorati after each a-posteriori filtering step.
\end{sloppypar}

\begin{table*}[t]
\begin{tabu}{X[l]|c|c|c|}
& \textbf{Precision} & \textbf{Recall} & \textbf{F-Measure} \\
\hline
&&&\\[-2ex]
BaseNEER & 13\% & 89\% & 0.23\\
BaseNEER + correlation filtering & 17\% & 74\% & 0.28 \\
BaseNEER + document frequency filtering & 50\% & 81\% & 0.62 \\
\end{tabu}
\caption{Evaluation results for BaseNEER applied to New York Times.}
\label{tab:results_baseneer}
\end{table*}

\begin{table*}[t]
\begin{tabu}{X[l]|c|c|c|}
& \textbf{Precision} & \textbf{Recall} & \textbf{F-Measure} \\
\hline
&&&\\[-2ex]
BlogNEER without a-posteriori filtering & 6\% & 64\% & 0.10 \rule{0pt}{2.5ex}\\
BlogNEER after a-posteriori frequency filtering & 48\% & 43\% & 0.45 \\
BlogNEER after semantic filtering & 67\% & 36\% & 0.47
\end{tabu}
\caption{Evaluation results for BaseNEER and BlogNEER applied to Blogs08.}
\label{tab:results_blogs08}
\end{table*}

\begin{table*}[t]
\begin{tabu}{X[l]|c|c|c|}
& \textbf{Precision} & \textbf{Recall} & \textbf{F-Measure} \\
\hline
&&&\\[-2ex]
BlogNEER without a-posteriori filtering & 6\% & 87\% & 0.11 \rule{0pt}{2.5ex}\\
BlogNEER after a-posteriori frequency filtering & 61\% & 77\% & 0.68 \\
BlogNEER after semantic filtering & 70\% & 67\% & 0.68
\end{tabu}
\caption{Evaluation results for BaseNEER and BlogNEER applied to Technorati.}
\label{tab:results_technorati}
\end{table*}

Our results indicate BlogNEER behaves differently on blog data from different domains. Although both datasets consist of blogs, we observed much less noise with the Technorati dataset, which consists of rather professional blogs and is specialized in certain categories. Blogs08 on the other hand consists of arbitrary, unspecialized and partly private blogs from the TREC dataset. This is reflected by the higher precision we achieved on Technorati and a loss of just 20\% recall by applying the a-posteriori filters. In order to achieve a similar precision for Blogs08 we had to filter out much more noise. As the noise was more frequent in the Blogs08 dataset, this led to correct co-references being filtered out as well, which resulted in a recall loss of 28\% compared to the unfiltered result.

Compared to BaseNEER, the recall we achieved with BlogNEER is similar, when applied to Technorati. On the less professional, more noisy dataset Blogs08 the initial, only a-priori filtered, as well as a-posteriori filtered recall is significantly lower than the recall of BaseNEER. However, for the precision we achieved values of around 70\% on both datasets. This is noticeably higher than the best precision of 50\% for BaseNEER with document frequency filtering. As precision is rated higher for our purpose (s. Sect.~\ref{sec:problem}) we consider our overall results better than BaseNEER. Considering that BaseNEER was not able to produce comparable results on blog data at all, this is an encouraging result.

\begin{sloppypar}
Table~\ref{tab:filtering_example} presents an example from the result of BlogNEER applied to Blogs08 for the query term \textit{Sean Combs} with change period 2005. The expected terms from our test set are \textit{Diddy} or \textit{P. Diddy} and \textit{Puff Daddy}. The table shows the result set after the different a-posteriori filtering steps together with the resulting precision and recall. With the semantic filter we were able to increase the precision to 100\% by filtering out noise. Unfortunately, \textit{Diddy} was filter out too, even though it is a correct co-reference. The reason for this is that the semantic filter disambiguated \textit{Diddy} to \textit{Diddy - Dirty Money}, a band of Sean Combs. As it is related to Sean Combs, he is mentioned very often in that resource. Due to this, it is considered to be most similar among all provided resources for \textit{Diddy} on DBpedia. As a band is an organization and not a person, it was filtered out as a valid co-reference. It should be noted that it would not have been possible to disambiguate the name correctly as DBpedia does not contain a resource for the term \textit{Diddy} that refers to an entity representing \textit{Sean Combs}. Therefore, it was the best disambiguation we could achieve with the given information. In case the term would have been completely unknown by DBpedia, we would have kept it in the result. This shows, without all needed information, the semantic filter can lead to worse results. However, as precision is considered more important than recall, it is a better decision to filter out such a term than keeping it, as it might be truly incorrect.
\end{sloppypar}

\begin{table*}[t]
\begin{tabu}{l|X[l]|c|c|}
\textbf{Step} & \textbf{Result set} & \textbf{Precision} & \textbf{Recall} \\
\hline
Unfiltered & Sean, Sean Penn, Penn, Combs, Diddy, York, Puff, Puff Daddy, Daddy, MTV, Video, Video Music Awards, Music Awards, Music, Award, Boy, Rock, Chris Rock, Chris, Bad, Rapper, James, October, Scott, Hampton, WHITE, Town, Johnson, BET, Simmons, Power, Grammy, Angelina Jolie, Angelina, Jolie, Summer, Post, America, Hip, Aug, Latino, King, Stefani, Carter, Boston, War, Lord, Red, Knight & 12\% & 100\%\\
\hline
Frequency Filtering & Sean, Sean Penn, Combs, Diddy, Puff Daddy, Video Music Awards & 67\% & 100\%\\
\hline
Semantic Filtering & Puff Daddy & 100\% & 50\%
\end{tabu}
\caption{BlogNEER example with a-posteriori filtering steps for the query term \textit{Sean Combs} on Blogs08.}
\label{tab:filtering_example}
\end{table*}

\section{Discussions}
\label{sec:discussions}

Named Entity Evolution Recognition on the Web is more complicated compared to NEER on traditional newspapers by the amount of noise, mainly caused by the dynamic language that is used on Blogs, for example. Our goal was to adapt the BaseNEER approach to be more robust to this noise. The intention of BlogNEER is to filter out noisy terms and thus, achieve results in recall and precision that are comparable to BaseNEER. Our evaluation shows that the proposed filtering mechanisms achieve encouraging results. The a-priori filters reduce noise prior to the co-reference detection while a-posteriori filters, including semantic filtering, increase the accuracy by filtering the results afterwards. Our evaluation of BaseNEER on Web data showed extremely low precision and proved the need for advanced and semantic filtering for this specific domain. With BlogNEER we achieved a significantly higher precision on Blogs and even obtained results which are comparable to BaseNEER applied to a newspaper dataset. The recall of BlogNEER compared to BaseNEER on the New York Times dataset is similar on Technorati, but lower on Blogs08.

\begin{sloppypar}
In this section we discuss how each extension on BaseNEER affected the results. We review the utilized methods and give ideas for future work on BlogNEER.
\end{sloppypar}

\subsection{Pre-processing and A-priori Filtering}

\begin{sloppypar}
The low recall achieved on Blogs08 is caused by the frequent noisy terms in the dataset, which we consider distinctive for the Web. The noise does not affect the recall directly, however, it leads to larger contexts. As these contexts contain many frequent terms that may be related to the query term (complementary terms) but are not co-references, they weaken the relative frequency of the actual co-references. By filtering out these frequent noisy terms using frequency filtering techniques, we filter out correct co-references of the query term as well. Thus, the a-priori frequency filter lowers the recall even before the terms reach the co-reference detection step. This leads to a lower recall of 64\% for BlogNEER, even before the a-posteriori filtering, compared to 90\% of BaseNEER. On the more qualitative Technorati dataset, which is not representative for the Web in general as it consists only of top blogs that are typically professional, yet still in ``Web language'', BlogNEER achieves a recall of 87\% before the a-posteriori filtering. This is close to the result of BaseNEER applied to the high quality New York Times dataset.
\end{sloppypar}

The dataset reduction step (s. Sect.~\ref{sec:dataset_reduction}) in the Blog\-NEER process helps to focus on the documents that are relevant for a query. With this step relevant terms are emphasized and become more frequent in relation to terms that are not related to the query. Consider the query \textit{President Obama} with the presidential election as its change period and imagine, for some reason, at the same time sport blogs extensively report about the president of some sports club. As the query is performed for the query sub-terms separately (i.e., \textit{President} and \textit{Obama}) the articles about the sports president would be considered as well as the articles about the presidential election. Thus, the name of the sport club and its president are most likely among the most frequent terms. By filtering out the documents from the sport blogs in the dataset reduction step, the frequency gap between the intended \textit{President Obama} and the sports president can be increased.

Our results indicate that pre-processing and a-priori filtering are a crucial parts in the NEER process. To overcome the challenges discussed above, further investigation is required to obtain a higher initial precision and recall. The recall of the context graph limits the recall of the entire NEER process and is therefore an important step. Clustering techniques could support differentiating between e.g., different domains and only retrieving documents from the domain of the query term. 

\subsection{A-posteriori Filtering}

In contrast to the a-priori filtering we could evaluate each a-posteriori filtering step using precision and recall (s. Sect.~\ref{sec:experiments}). The evaluation showed that both the a-posteriori frequency as well as the semantic filters are very effective in increasing precision. As is mostly the case, high precision comes at the expense of recall (after frequency filtering we have a decrease in recall by 21\% on Blogs08 and 10\% on Technorati, after semantic filtering by 28\% on Blogs08 and 20\% on Technorati). Even though the semantic filter sometimes filters out correct co-references erroneously, it is more effective in filtering false positives. Hence, for queries for which BlogNEER does not detect any real co-references, the semantic filter can still filter out most false positive candidates. This indicates, BlogNEER with semantic filtering can, in addition to finding co-references with a high precision, able to filter out false positive co-references for terms that have not changed. 

The semantic filtering proposed with BlogNEER is the first approach to involve external resources in the BaseNEER method. Because of the wide-spread use of the Web and the search mechanism for finding information on the Web, coupled with the increasing time spans of the documents, there is a need for reliable NEER detection in this domain. Therefore, the next logical step for NEER is to make use of Web data. In this paper, we have taken a first step in this direction.  As our evaluation results show, with semantic filtering we achieved a precision gain of 19\% while losing only 7\% recall on Blogs08 compared to the result after a-posteriori frequency filtering. On Technorati the precision gain with the semantic filtering is 9\% with a recall loose of 10\%. However, the semantic filter currently has one decisive drawback; it can only be applied to terms that are known by a knowledge base. However, for these terms, the co-references are typically known as well. This issue is discussed in the following subsection.

\subsubsection{Redundant Information and Unknown Terms}
\label{sec:semantic_redundancy}

In a separate experiment that we conducted to check the terms from our test set for resources on DBpedia, we found that nine out of 30 entities with co-references actually point to different resources. An example is \textit{Czechoslovakia}. Although the country split into two separate countries, its name evolved from the old to the new names, still describing the same area, even if politically not the same country. Thus, from the perspective of NEER, this is a name evolution and important for a search engine to be aware of. However, on DBpedia, \textit{Czechoslovakia} as well as both countries that it was split to (\textit{Czech Republic} and \textit{Slovakia}) have their own resources, not including a hint to the other names. For the remaining 21 entities, all co-references that could be found on DBpedia redirected to the same resource, which means the co-references were already known. The semantic filter does not use this information explicitly, but it uses it implicitly when it compares the corresponding resources, which are the same in these cases. Therefore, it considers those co-references as correct because they are not just similar but exactly equal. In this case, NEER would not have been needed as the co-references are already available on DBpedia and thus, redundant. However, we also found nine out of the 21 remaining entities with co-references that are not known on DBpedia. These might be detected by NEER and indeed would not be filtered out, because unknown terms are not semantically comparable. While they are not filtered out as incorrect, however, they are not proven to be correct either. And as soon the terms become available on DBpedia, there will most likely exist a redirection from or to the other co-references, too.

\begin{sloppypar}
Still, the semantic filter improves the recall very effectively, as shown by our evaluation, because it can filter out terms that definitely do not refer to the queried entity. For example, \textit{Microsoft}, a company, can be filtered out as co-reference for \textit{Kinect}, a gaming device. This works because both terms are known on DBpedia. Unknown terms can not be filtered with the semantic filter as proposed in this article. While those terms might be co-references, they could also be completely unrelated. In case the query term can not be resolved to a semantic resource, none of the found co-reference candidates can be filtered out as we cannot apply semantic filtering at all. Therefore, it remains a crucial issue for future work to recognize the type of an unknown term or in other ways to obtain  semantic information about unknown terms. One possibility can be to resolve just sub-terms instead of the whole term, for example \textit{Chancellor Angela Merkel} is not known on DBpedia while \textit{Angela Merkel} is.
\end{sloppypar}

\subsubsection{Entities of Same Type}

Another issue we faced was that with the semantic filter that depends on the similarity of two terms that denote entities of the same type. The semantic filter works very reliable on entities of different types, for example \textit{Pope Benedict} and the often co-occurring \textit{Vatican}. In this case, the filter identifies \textit{Pope Benedict} as a person and \textit{Vatican} as a place. Hence, it filters out \textit{Vatican} as co-reference for \textit{Pope Benedict}. However, if two terms are of the same kind, filtering is a much harder problem. For instance, \textit{Angela Merkel}, the current German chancellor, and \textit{Gerhard Schroeder}, the former German chancellor, are both of the type person. As one is the successor of the other, they have at least one year in common, the election when \textit{Angela Merkel} became new chancellor. Additionally, both are in the category \textit{Chancellors of Germany}. Therefore, with the proposed semantic filter we were not able to tell them apart. The same problem happens, for example, for countries of the same form of government, like republics. For instance, \textit{China} was not filtered out as co-reference of \textit{Burma}.

A possible solution also in this case would be to compare just sub-terms with each other. With the example above, a reasonable comparison would be to compare the sub-terms of the same type. For instance, \textit{Angela} and \textit{Gerhard} are both first names. Comparing these, we would have been able to tell them apart: \textit{Gerhard} is in the category of masculine given names while \textit{Angela} is in the category of feminine given names. It might also be possible to include the sub-term information as an extra element in the total score of the filter, however, the method would not work for \textit{China} and \textit{Burma}. 

Another way would be to incorporate type specific semantic properties. So far, we only used general information that are common among all resource classes on DBpedia, like subjects and types. However, by identifying \textit{Angela Merkel} and \textit{Gerhard Schroeder} both as persons, we could compare their birth dates for example. Thus, we would have been able to identify them as different persons. As birth date is a specific property just for persons, this would not work for comparing cities. Hence future work in this direction would be to tailor fit properties for different kind of entities which requires manual work from domain experts.

\subsubsection{Resolving Names and Disambiguation}

\begin{sloppypar}
Correctly resolving of the query term and its co-reference candidates on DBpedia as well as the disambiguating in cases where the terms have different meanings is a crucial step for the semantic filtering, as the example of \textit{Sean Combs} in Sect.~\ref{sec:results} showed. In early experiments we had problems because many terms were disambiguated to the wrong resource. For instance, disambiguating \textit{Pope Benedict} to \textit{Pope Benedict III} instead of \textit{Pope Benedict XVI} leads to wrong results as it will be filtered out as co-reference for \textit{Joseph Ratzinger}. However, we have made good progress in this area and with the strategies presented in Sect.~\ref{sec:resolving_of_names} and Sect.~\ref{sec:disambiguation} we achieved correct resolved resources for almost all tested terms. However, as the \textit{Sean Combs} example shows, the method only works if the intended disambiguation entity is available on DBpedia.
\end{sloppypar}

\begin{sloppypar}
In order to resolve a term to its semantic resource we start with the term itself. In case this term does not exist on DBpedia we proceed with its direct co-reference, as explained in Sect.~\ref{sec:resolving_of_names}. Afterwards, in case the resolved term has multiple meanings on DBpedia, it is disambiguated using two strategies. The direct disambiguation incorporates the direct co-references that have been found for the term. The indirect disambiguation utilizes cosine similarity, calculated between the DBpedia resource and the indirect co-reference candidates of the query term. If a term cannot be disambiguated directly we select the most similar resource for the semantic filtering. However, this only works correctly if the intended resource is available on DBpedia at all. Otherwise, we take the most similar one, which is still incorrect though. An example is the American football player \textit{Chad Johnson}, also known as \textit{Ocho Cinco}. While \textit{Ocho Cinco} points to the right resource there is no disambiguation for \textit{Chad Johnson} as football player. Instead, our semantic filter disambiguates the term to \textit{Chad Johnson (ice hockey)} and filters out \textit{Ocho Cinco} as an incorrect co-reference. Interestingly, the disambiguation for \textit{Chad Johnson} as football player is already available on Wikipedia, it has just not been updated on DBpedia yet.  
\end{sloppypar}

\begin{sloppypar}
This issue might be solvable with a more complete knowledge base. Alternatives to DBpedia are among others Yago\footnote{\url{http://www.yago-knowledge.org}} or Freebase\footnote{\url{http://www.freebase.com}}. However, no database can include all entities. Therefore, future work needs to investigate how to decide that terms are not the same (i.e., sufficiently dissimilar), like the intended \textit{Chad Johnson} and the ice hockey player.
\end{sloppypar}

\section{State of the Art}
\label{sec:sota}

The analysis of cultural trends in big collections of digital texts has lately become known as Culturomics, introduced by \citet{Michel2010}. This field of research deals with the detection of cultural trends by analyzing the use of language. Hence, named entity evolution recognition, a discipline of language evolution detection, can be regarded as an effort of Culturomics.

\begin{sloppypar}
Just like NEER, most previous work on automatic language evolution detection has
mainly focused on named entity evolution. Still, the number of published work in this area is very limited. In the available papers, the interest is mainly motivated by information retrieval (IR). For IR the awareness of terminology evolution is highly relevant. In order to overcome the word mismatch problem caused by texts from different times, search engines depend on temporal information. Especially on data sources that cover a large timespan, like digital archives, the knowledge of termporal co-references affects the effectiveness of a search engine. In contrast to commonly used \textit{co-references}, which are ``'equivalent' URIs referring to the same concept or entity'' \cite{Glaser2009}, we consider \textit{temporal co-references} as terms that refer to the same entity. These may or may not be emerged in different periods of time.
\end{sloppypar}

In the following we present different works on language evolution in the fields of word sense evolution as well as term to term evolution with alternative approaches to language evolution detection. Afterwards, we will walk down to the foundations of BlogNEER, our NEER approach that has been presented in this article, to give an overview of the related fields.

\subsubsection*{Word Sense Evolution} 
Automatic detection of changes and variations in word senses over time is a topic that is increasingly gaining interest. During the past years researchers have evaluated and researched different parts of the problem mainly in the field of computational linguistics. 

\citet{Sagi2009} presented work on finding \textit{narrowing} and \textit{broadening} of senses over time by applying semantic density analysis. Their work provides indication of semantic change, unfortunately without  clues to what has changed but can be used as an initial warning system. 

The work presented by \citet{Lau2012} aims to detect word senses that are novel in a later corpus compared to an earlier one and use LDA topics to represent word senses. Overall, the method shows promising results for detecting novel (or outdated) word senses by means of topic modeling. However, alignment of word senses over time or relations between senses is not covered in this work.

\begin{sloppypar}
\citet{Wijaya2011} report on automatic tracking of word senses over time by  clustering topics. Change in meaning of a term is assumed to correspond to a change in cluster for the corresponding topic. A few different words are analyzed and there is indication that the method works and can find periods when words change their primary meaning. In general, the work in this paper is preliminary but with promising indications.
\end{sloppypar}

\begin{sloppypar}
A previous work presented by \citet{Tahmasebi2013} was the first to automatically track individual word senses over time to determine changes in the meanings of terms. It shows \textit{narrowing} and \textit{broadening} as well as slow shifts in meaning in individual senses and relations between senses over time like \textit{splitting}, \textit{merging}, \textit{polysemy} and \textit{homonymy}. For most of the evaluated terms, the automatically extracted results corresponded well to the expected evolution. However, word senses were not assigned to individual word instances, which is necessary to help users understand individual documents. 
\end{sloppypar}

\subsection{Term to Term Evolution}

\begin{sloppypar}
\citet{Berberich2009} proposed a query reformulation technique to translate a queries into terms used in older texts. Therefore, they consider the entire query phrase, not just single query terms. Moreover, other than replacing names of entities with their temporal co-references, they replace terms referring to a concept with older terms referring to the same concept (e.g. \textit{walkman} and \textit{ipod}). The detection of these terms is based on a Hidden Markov Model using the relatedness among two terms. The relatedness is computed by means of the contexts of the terms. Similarly to us, they consider a context of a term as frequently co-occurring terms. However, this approach needs a recurring processing for each query at the cost of performance.
\end{sloppypar}

\citet{Kaluarachchi2010} approached the performance drawback by pre-computing temporally related terms. This has been achieved by using a machine learning technique to mine association rules based on events, corresponding to verbs. The nouns that are referred to by similar verbs are considered semantically related. By incorporating the time stamps of the events, these related terms are used for a temporal query translation. The method could be used for shorter time spans but is less suited for longer time spans as verbs are more likely to change over time than nouns \cite{Sagi2010}.

In a more explicit way \citet{Kanhabua2010} discover temporally related terms by using the history of Wikipedia. They extract anchor texts from articles of Wikipedia snapshots at different times. Those texts that point to the same entity page are considered time-based synonyms of that entity. This approach might be well suites for evolutions that are well known, however, it does not work for name changes mentioned in texts, which have not been extracted and stored in a knowledge base yet.
 
\begin{sloppypar}
In more recent work, \citet{Mazeika2011} presented a tool for analyzing named entity evolution by means of entity timelines. Rather than automatically detecting the name changes though, the timelines visualize the evolution of named entities together with other co-occurring entities. For entity extraction and disambiguation they incorporate the YAGO knowledge base (compare Sect.~\ref{sec:foundations}).
\end{sloppypar}

\begin{sloppypar}
Most related to our work on BlogNEER is the NEER approach by \citeauthor{Tahmasebi2013} \cite{Tahmasebi2012, Tahmasebi2013}, referred to as BaseNEER in this article. It is an unsupervised method for named entity evolution recognition in a high quality newspaper (i.e., New York Times). Using \textit{change periods}, the method avoid comparing terms from arbitrary time periods and thus overcome a severe limitation of existing methods; the need to compare co-occurring terms or associated events from different time periods. The approach is described in more detail in Sect.~\ref{sec:NEER}. BaseNEER builds the foundation of BlogNEER and originated this work. We direct it towards the Web by improving its resistance against noise and incorporating external semantic resources.
\end{sloppypar}

Together with the work on BaseNEER we developed \textit{fokas}, a search engine to demonstrate the potential of NEER in supporting IR. From the IR perspective, NEER can be regarded as a query expansion method. Query expansion is the task of finding additional terms, like synonyms or related concepts, to automatically expand a query for matching texts that use a different vocabulary \cite{Voorhees1994}. The temporal co-references detected by NEER can be used to expand the initial query in order to find documents from other periods of time. \textit{Fokas} makes use of NEER for query expansion. The description of the demo has been published in \citet{fokas}. \textit{Fokas} is a simple search engine with typical look and feel that incorporates temporal co-references derived by BaseNEER. For each query \textit{fokas} presents the detected temporal co-references to the user along with a chart that shows the frequencies of these terms over time. By selecting one or more of the co-reference terms, \textit{fokas} automatically reformulates the query to find texts that contain these terms, too. Search results that have been found as a result of query expansion with temporal co-references are marked accordingly.

During the work on BlogNEER, the preliminary results were published in \citet{Holzmann2013}. A more comprehensive description of this work can be found in the corresponding Master's thesis by \citet{Holzmann2013a}.

\subsection{Foundations of BlogNEER}
\label{sec:foundations}

\begin{sloppypar}
BlogNEER is our approach to NEER on the Web. Besides operating on Web data it also incorporates the Semantic Web to identify erroneously detected temporal co-references. This semantic filtering raises new problems from the areas of entity linking and entity resolution, semantic similarity measuring as well as word sense disambiguation. According to \citet{Mendes2012} knowledge bases, like DBpedia, can support most of these tasks by providing semantic data. A knowledge base is a collection of structured data based on an ontology. The result of an early effort to create such database with lexical and semantic information for the English language is WordNet \cite{Miller1995}. While WordNet is maintained by a closed group of experts, other knowledge bases are a community effort. The data on DBpedia and Yago has mainly been extracted from Wikipedia \cite{Bizer2009, Suchanek2007}. Commonly used sources are for example infoboxes in Wikipedia articles as well as links to other entities. Additionally, Yago incorporates the class hierarchy of WordNet to classify its entities. This information are in turn used by DBpedia since both knowledge bases are interlinked among each other. Freebase, another knowledge base, employs Wikipedia too \cite{Bollacker2008}. Additionally, users can edit the database directly. Therefore, other than Yago and DBpedia it does not necessarily rely on updates from Wikipedia.
\end{sloppypar}

DBpedia spotlight by \citet{Mendes2011} is a tool to automatically annotate named entities in texts with data from DBpedia. Since BlogNEER also uses DBpedia as a knowledge base for semantic filtering we need to perform similar tasks. These can be summarized under entity linking: ``The Entity Linking task requires aligning a textual mention of a named-entity (a person, organization, or geo-political entity) to its appropriate entry in the knowledge base, or correctly determining that the entity does not have an entry in the KB'' \citep{McNamee2010}. One of the core sub-tasks of entity linking is word sense disambiguation, which is a crucial part in our work too. ``Word sense disambiguation is the process of assigning a meaning to a word based on the context in which it occurs'' \cite{Patwardhan2003}.

\citet{Banerjee2002} propose a disambiguation by means of WordNet. Similar to the early disambiguation algorithm by \citet{Lesk1986} they take the surrounding words into account and measure their relatedness to different senses of the ambiguous term. Other approaches in this area work with different knowledge sources. Instead of words in a text, \citet{Garca2009} link tags to entries on DBpedia. Tags are separate terms that describe an article. The disambiguation of a tag is based on the similarity between co-occurring tags and a DBpedia entry, both represented as term vectors. The similarity is computed by means of cosine similarity. The most similar entry provided by DBpedia as disambiguation candidate is taken as a semantic representation of the tag.

After the entity linking we need to compute the similarity among resources of two terms in order to decide whether or not these refer to the same entity. This task is related to entity resolution, which tackles the problem of identifying entity representations that refer to the same entity. A famous technique for entity resolution is blocking \cite{Whang2009}. Blocking groups similar lexical references, like database records or names, to blocks of candidates and compares these in order to determine which of them refer to the same entity. We perform a similar task to decide among a group of temporal co-references if the corresponding DBpedia entries refer to the same entity. Similarly to the semantic type filtering of BlogNEER, an early work by \citet{Richardson1994} already analyzed the class hierarchy of a concept to compute the similarity among two concepts. They used again WordNet since DBpedia was not available at that time.

\section{Conclusions and Outlook}
\label{sec:conclusions}

Language evolves over time. This leads to a gap between language known to the user and language used in documents stored in digital archives. To ensure that the content in our digital libraries can be found and semantically interpreted, we must consider \textbf{semantic preservation} and prepare our archives for future processing and long-term storage. Automatic detection of language evolution is a first step towards offering semantic access, however, several other measures need to be taken. Dictionaries, natural language processing tools and other resources must be stored alongside each archive to help processing in the future. Data structures and indexes that respect temporal evolution are needed to utilize language evolution for searching, browsing and understanding of content. To take full advantage of continuously updated archives that do not require expensive, full re-computation with each update, we must invest effort into transforming our digital archives into \textbf{living archives} that continuously learn about changes in language. 

In this article we presented BlogNEER, an approach towards Named Entity Evolution Recognition (NEER) on the Web and the Blogosphere in particular. Texts on the Web are different from texts in traditional newspapers. They are written in a more dynamic language as people can express their thoughts in their every day language, even in written texts and still be published online. There is no need to follow rules in writing. All terms from spoken language can be used. Additionally, many complementary terms are mentioned in order to be linked in an article. This diversity of terms means more noise for the NEER task and makes it difficult to identify the actual temporal co-references of a term without taking the noisy terms as well. At the same time, this new context provides new opportunities.

\begin{sloppypar}
BlogNEER extends an existing unsupervised approach to NEER for temporal co-reference detection on high quality newspaper texts, referred to as BaseNEER. An additional pre-processing step that reduces the dataset with respect to the query term helps to focus on the relevant documents. Advanced frequency filters, a-priori and a-posteriori, reduce the number of derived co-reference candidates in the result set. Moreover, BlogNEER is the first NEER approach that exploits external resources from the Semantic Web to filter results found using the text itself. This turns out to have a big potential. Incorporating semantic properties, such as types and subjects (i.e., categories), help to tell apart terms that refer to different entities. With this information we are able to identify names that can not constitute name evolutions of each other.
\end{sloppypar}

Our proposed methods achieved a similar recall on Web data as BaseNEER on a traditional newspaper dataset. However, we observed big deviations for different kinds of data. On professional blogs we achieved a higher initial recall of 87\% and were able to keep it relatively high at 67\% by increasing precision with filtering. On a second dataset, consisting of rather random blogs, we could only achieve a recall of 36\%, which is lower than BaseNEER's recall on newspaper data of 81\%. However, on both datasets we achieved a good precision of around 70\%, which is noticeably higher than the best precision of BaseNEER with document frequency filtering of 50\%. These results are highly encouraging and show that NEER using the BlogNEER approach can be performed on blog data as well. Even though the recall may be lower due to noise the stricter filtering required, we are still able to produce good results.

\begin{sloppypar}
The largest drawback of the semantic filter is its dependency on the existence of resources in semantic knowledge bases for the terms under consideration. Incorporating semantic information of the sub-terms, such as first and last names or titles of persons, might help to overcome this issue. It would allow to use meta data of a name, like the gender derived by the first name of an entity. Hence, the full name does not need to exist in a knowledge base to be able to reduce noise.
\end{sloppypar}

\begin{sloppypar}
Our method for NEER was used in the same fashion on all terms in the test set. No distinction was made between terms for which the name changes were documented in DBpedia compared to those without documented changes. In future work, we will investigate a combination of methods where known name changes are taken directly and unknown changes are found using NEER. NEER can be applied only where names, redirects and other information is not available or as a complement to existing names. Also, already known name changes can be used to find additional name changes not documented in knowledge-resources. These additional names can include informal nicknames used by the public.

\end{sloppypar}

\begin{sloppypar}
 In addition, we will investigate automatic detection of change periods on Web data. In this work we used the change periods from the BaseNEER test set. In order to perform NEER independently on Web data, we must be able to find change periods on the Web in an automatic fashion. Ideally, this will take place live while the data is published. Therefore, we need to investigate NEER on streams with a burst-detection-like method in order to detect change periods as early as possible. These properties will allow immediate evolution detection as a step forward towards \textit{living} semantic digital archives.
\end{sloppypar}

\renewcommand\bibname{References}
\bibliography{bib}{}

\begin{thebibliography}{41}
\providecommand{\natexlab}[1]{#1}
\providecommand{\url}[1]{\texttt{#1}}
\expandafter\ifx\csname urlstyle\endcsname\relax
  \providecommand{\doi}[1]{doi: #1}\else
  \providecommand{\doi}{doi: \begingroup \urlstyle{rm}\Url}\fi

\bibitem[Segerstad(2002)]{segerstad2002use}
Y.H. Segerstad.
\newblock \emph{Use and adaptation of written language to the conditions of
  computer-mediated communication}.
\newblock PhD thesis, University of Gothenburg, 2002.

\bibitem[Tahmasebi et~al.(2012{\natexlab{a}})Tahmasebi, Gossen, Kanhabua,
  Holzmann, and Risse]{Tahmasebi2012}
Nina Tahmasebi, Gerhard Gossen, Nattiya Kanhabua, Helge Holzmann, and Thomas
  Risse.
\newblock Neer: An unsupervised method for named entity evolution recognition.
\newblock In Martin Kay and Christian Boitet, editors, \emph{Proceedings of the
  24th International Conference on Computational Linguistics (Coling 2012)},
  pages 2553--2568, Mumbai, India, December 2012{\natexlab{a}}. Indian
  Institute of Technology Bombay.
\newblock URL \url{http://www.l3s.de/neer-dataset}.

\bibitem[Tahmasebi et~al.(2012{\natexlab{b}})Tahmasebi, Gossen, and
  Risse]{tpdl2012}
Nina Tahmasebi, Gerhard Gossen, and Thomas Risse.
\newblock Which words do you remember? temporal properties of language use in
  digital archives.
\newblock In \emph{Theory and Practice of Digital Libraries}, volume 7489,
  pages 32--37. Springer, 2012{\natexlab{b}}.

\bibitem[The~{T}imes(1942)]{Times-Stalingrad}
DIPLOMATIC~CORRESPONDENT The~{T}imes.
\newblock {Menace To The Volga.}
\newblock In \emph{London, England, Jul 17, 1942; pg. 3; Issue 49290.} Gale
  Doc. No.: CS52116209, 1942.

\bibitem[{The Times}(1787)]{Times-Opera}
{The Times}.
\newblock {Sestini's benefit last night at the Opera-House was overflowing with
  the fashionable and gay.}
\newblock In \emph{London, England, Apr 27, 1787; pg. 3; Issue 736.} Gale Doc.
  No.: CS50726043, 1787.

\bibitem[Berberich et~al.(2009)Berberich, Bedathur, Sozio, and
  Weikum]{Berberich2009}
Klaus Berberich, Srikanta~J. Bedathur, Mauro Sozio, and Gerhard Weikum.
\newblock Bridging the terminology gap in web archive search.
\newblock In \emph{WebDB}, 2009.

\bibitem[Bizer et~al.(2009)Bizer, Lehmann, Kobilarov, Auer, Becker, Cyganiak,
  and Hellmann]{Bizer2009}
Christian Bizer, Jens Lehmann, Georgi Kobilarov, S\"{o}ren Auer, Christian
  Becker, Richard Cyganiak, and Sebastian Hellmann.
\newblock {DBpedia - A crystallization point for the Web of Data}.
\newblock \emph{Journal of Web Semantics}, 7\penalty0 (3):\penalty0 154--165,
  September 2009.
\newblock ISSN 1570-8268.
\newblock \doi{10.1016/j.websem.2009.07.002}.
\newblock URL \url{http://dx.doi.org/10.1016/j.websem.2009.07.002}.

\bibitem[Garc{\'{a}}-Silva et~al.(2009)Garc{\'{a}}-Silva, Szomszor, Alani, and
  Corcho]{Garca2009}
A.~Garc{\'{a}}-Silva, M.~Szomszor, H.~Alani, and O.~Corcho.
\newblock {Preliminary results in tag disambiguation using DBpedia}.
\newblock In \emph{Knowledge Capture (K-Cap 2009)-Workshop on Collective
  Knowledge Capturing and Representation-CKCaR}, 2009.

\bibitem[Inc.(2013)]{technorati}
Technorati Inc.
\newblock {accessed June 05, 2013}, 2013.
\newblock URL \url{http://www.technorati.com}.

\bibitem[Ounis et~al.(2009)Ounis, Macdonald, and Soboroff]{blogs08}
Iadh Ounis, Craig Macdonald, and Ian Soboroff.
\newblock Overview of the trec-2008 blog track.
\newblock In \emph{In Proceedings of TREC-2008}, 2009.

\bibitem[Tahmasebi et~al.(2012{\natexlab{c}})Tahmasebi, Gossen, Kanhabua,
  Holzmann, and Risse]{testset}
Nina Tahmasebi, Gerhard Gossen, Nattiya Kanhabua, Helge Holzmann, and Thomas
  Risse.
\newblock Named entitiy evolution dataset.
\newblock Available online at \url{http://l3s.de/neer-dataset},
  2012{\natexlab{c}}.

\bibitem[Cornolti et~al.(2013)Cornolti, Ferragina, and Ciaramita]{Cornolti2013}
Marco Cornolti, Paolo Ferragina, and Massimiliano Ciaramita.
\newblock A framework for benchmarking entity-annotation systems.
\newblock In \emph{WWW}, pages 249--260, 2013.

\bibitem[Mendes et~al.(2011)Mendes, Jakob, García-Silva, and Bizer]{Mendes2011}
Pablo~N. Mendes, Max Jakob, Andrés García-Silva, and Christian Bizer.
\newblock Dbpedia spotlight: shedding light on the web of documents.
\newblock In \emph{I-SEMANTICS}, pages 1--8, 2011.

\bibitem[Rails-Core-Team()]{ruby_on_rails}
Rails-Core-Team.
\newblock Ruby on rails.
\newblock (accessed August 29, 2013).
\newblock URL \url{http://rubyonrails.org}.

\bibitem[Finkel et~al.(2005)Finkel, Grenager, and Manning]{StanfordNER}
Jenny~Rose Finkel, Trond Grenager, and Christopher Manning.
\newblock Incorporating non-local information into information extraction
  systems by {G}ibbs sampling.
\newblock In \emph{ACL}, pages 363--370, 2005.

\bibitem[Kleinberg(2003)]{Kleinberg2003}
Jon~M. Kleinberg.
\newblock Bursty and hierarchical structure in streams.
\newblock \emph{Data Min. Knowl. Discov.}, 7\penalty0 (4):\penalty0 373--397,
  2003.

\bibitem[Porter(1980)]{Porter1997}
Martin~F Porter.
\newblock An algorithm for suffix stripping.
\newblock \emph{Program: electronic library and information systems},
  14\penalty0 (3):\penalty0 130--137, 1980.

\bibitem[Holzmann(2013)]{Holzmann2013a}
Helge Holzmann.
\newblock Webneer: Towards named entity evolution recognition on the web.
\newblock Master's thesis, University of Hanover, L3S Research Center, 2013.

\bibitem[Michel et~al.(2010)Michel, Shen, Aiden, Veres, Gray, Team, Pickett,
  Holberg, Clancy, Norvig, Orwant, Pinker, Nowak, and Aiden]{Michel2010}
Jean-Baptiste Michel, Yuan~Kui Shen, Aviva~Presser Aiden, Adrian Veres,
  Matthew~K. Gray, The Google~Books Team, Joseph~P. Pickett, Dale Holberg, Dan
  Clancy, Peter Norvig, Jon Orwant, Steven Pinker, Martin~A. Nowak, and
  Erez~Lieberman Aiden.
\newblock Quantitative analysis of culture using millions of digitized books.
\newblock \emph{Science}, 2010.

\bibitem[Glaser et~al.(2009)Glaser, Jaffri, and Millard]{Glaser2009}
Hugh Glaser, Afraz Jaffri, and Ian Millard.
\newblock Managing co-reference on the semantic web.
\newblock In \emph{LDOW}, 2009.

\bibitem[Sagi et~al.(2009)Sagi, Kaufmann, and Clark]{Sagi2009}
Eyal Sagi, Stefan Kaufmann, and Brady Clark.
\newblock Semantic density analysis: comparing word meaning across time and
  phonetic space.
\newblock In \emph{Proc. of the Workshop on Geometrical Models of Natural
  Language Semantics}, GEMS '09, pages 104--111. ACL, 2009.

\bibitem[Lau et~al.(2012)Lau, Cook, McCarthy, Newman, and Baldwin]{Lau2012}
Jey~Han Lau, Paul Cook, Diana McCarthy, David Newman, and Timothy Baldwin.
\newblock {Word Sense Induction for Novel Sense Detection}.
\newblock In \emph{EACL}, pages 591--601, 2012.

\bibitem[Wijaya and Yeniterzi(2011)]{Wijaya2011}
Derry~Tanti Wijaya and Reyyan Yeniterzi.
\newblock Understanding semantic change of words over centuries.
\newblock In \emph{Proceedings of the Int. workshop on DETecting and Exploiting
  Cultural diversiTy on the social web}, DETECT '11, pages 35--40. ACM, 2011.

\bibitem[Tahmasebi(2013)]{Tahmasebi2013}
Nina Tahmasebi.
\newblock \emph{{Models and Algorithms for Automatic Detection of Language
  Evolution. Towards Finding and Interpreting of Content in Long-Term
  Archives}}.
\newblock PhD thesis, Leibniz Universit{\"a}t Hannover, 2013.

\bibitem[Kaluarachchi et~al.(2010)Kaluarachchi, Varde, Bedathur, Weikum, Peng,
  and Feldman]{Kaluarachchi2010}
Amal~Chaminda Kaluarachchi, Aparna~S. Varde, Srikanta~J. Bedathur, Gerhard
  Weikum, Jing Peng, and Anna Feldman.
\newblock Incorporating terminology evolution for query translation in text
  retrieval with association rules.
\newblock In \emph{CIKM}, pages 1789--1792. ACM, 2010.

\bibitem[Sagi(2010)]{Sagi2010}
Eyal Sagi.
\newblock {Nouns are more stable than Verbs: Patterns of semantic change in
  19th century English}.
\newblock \emph{The 32nd Annual Conference of the Cognitive Science Society},
  2010.

\bibitem[Kanhabua and N{\o}rv{\aa}g(2010)]{Kanhabua2010}
Nattiya Kanhabua and Kjetil N{\o}rv{\aa}g.
\newblock Exploiting time-based synonyms in searching document archives.
\newblock In \emph{Proceedings of the 10\textsuperscript{th} annual joint
  conference on Digital libraries}, JCDL '10, pages 79--88, New York, NY, USA,
  2010. ACM.

\bibitem[Mazeika et~al.(2011)Mazeika, Tylenda, and Weikum]{Mazeika2011}
Arturas Mazeika, Tomasz Tylenda, and Gerhard Weikum.
\newblock Entity timelines: visual analytics and named entity evolution.
\newblock In \emph{CIKM}, pages 2585--2588, 2011.

\bibitem[Voorhees(1994)]{Voorhees1994}
Ellen~M. Voorhees.
\newblock Query expansion using lexical-semantic relations.
\newblock In \emph{SIGIR}, pages 61--69, 1994.

\bibitem[Holzmann et~al.(2012)Holzmann, Gossen, and Tahmasebi]{fokas}
Helge Holzmann, Gerhard Gossen, and Nina Tahmasebi.
\newblock fokas: Formerly known as - a search engine incorporating named entity
  evolution.
\newblock In Martin Kay and Christian Boitet, editors, \emph{Proceedings of the
  24th International Conference on Computational Linguistics: Demonstration
  Papers (Coling 2012)}, pages 215--222, Mumbai, India, December 2012. Indian
  Institute of Technology Bombay.
\newblock URL \url{http://www.l3s.de/neer-dataset/fokas.html}.

\bibitem[Holzmann et~al.(2013)Holzmann, Tahmasebi, and Risse]{Holzmann2013}
Helge Holzmann, Nina Tahmasebi, and Thomas Risse.
\newblock Blogneer: Applying named entity evolution recognition on the
  blogosphere.
\newblock In \emph{In Proceedings of the 3rd International Workshop on Semantic
  Digital Archives (SDA 2013), in conjunction with the 17th International
  Conference on Theory and Practice of Digital Libraries (TPDL 2013)}, Valetta,
  Malta, September 2013.

\bibitem[Mendes et~al.(2012)Mendes, Jakob, and Bizer]{Mendes2012}
Pablo~N. Mendes, Max Jakob, and Christian Bizer.
\newblock Dbpedia: A multilingual cross-domain knowledge base.
\newblock In \emph{LREC}, pages 1813--1817, 2012.

\bibitem[Miller(1995)]{Miller1995}
George~A. Miller.
\newblock Wordnet: A lexical database for english.
\newblock pages 39--41, 1995.

\bibitem[Suchanek et~al.(2007)Suchanek, Kasneci, and Weikum]{Suchanek2007}
Fabian~M. Suchanek, Gjergji Kasneci, and Gerhard Weikum.
\newblock Yago: a core of semantic knowledge.
\newblock In \emph{WWW}, pages 697--706, 2007.

\bibitem[Bollacker et~al.(2008)Bollacker, Evans, Paritosh, Sturge, and
  Taylor]{Bollacker2008}
Kurt~D. Bollacker, Colin Evans, Praveen Paritosh, Tim Sturge, and Jamie Taylor.
\newblock Freebase: a collaboratively created graph database for structuring
  human knowledge.
\newblock In \emph{SIGMOD Conference}, pages 1247--1250, 2008.

\bibitem[McNamee et~al.(2010)McNamee, Dang, Simpson, Schone, and
  Strassel]{McNamee2010}
Paul McNamee, Hoa~Trang Dang, Heather Simpson, Patrick Schone, and Stephanie
  Strassel.
\newblock An evaluation of technologies for knowledge base population.
\newblock In \emph{LREC}, 2010.

\bibitem[Patwardhan et~al.(2003)Patwardhan, Banerjee, and
  Pedersen]{Patwardhan2003}
Siddharth Patwardhan, Satanjeev Banerjee, and Ted Pedersen.
\newblock Using measures of semantic relatedness for word sense disambiguation.
\newblock In \emph{CICLing}, pages 241--257, 2003.

\bibitem[Banerjee and Pedersen(2002)]{Banerjee2002}
Satanjeev Banerjee and Ted Pedersen.
\newblock An adapted lesk algorithm for word sense disambiguation using
  wordnet.
\newblock In \emph{CICLing}, pages 136--145, 2002.

\bibitem[Lesk(1986)]{Lesk1986}
Michael Lesk.
\newblock Automatic sense disambiguation using machine readable dictionaries:
  how to tell a pine cone from an ice cream cone.
\newblock In \emph{Proceedings of the 5th annual international conference on
  Systems documentation}, SIGDOC '86, pages 24--26, New York, NY, USA, 1986.
  ACM.
\newblock ISBN 0-89791-224-1.

\bibitem[Whang et~al.(2009)Whang, Menestrina, Koutrika, Theobald, and
  Garcia-Molina]{Whang2009}
Steven~Euijong Whang, David Menestrina, Georgia Koutrika, Martin Theobald, and
  Hector Garcia-Molina.
\newblock Entity resolution with iterative blocking.
\newblock In \emph{SIGMOD Conference}, pages 219--232, 2009.

\bibitem[Richardson et~al.(1994)Richardson, Smeaton, and
  Murphy]{Richardson1994}
R.~Richardson, A.~F. Smeaton, and J.~Murphy.
\newblock Using wordnet as a knowledge base for measuring semantic similarity
  between words.
\newblock Technical report, In Proceedings of AICS Conference, 1994.

\end{thebibliography}
\bibliographystyle{unsrtnat}

\end{document}